\documentclass[11pt,a4paper]{article}
\usepackage[utf8]{inputenc}
\usepackage[T1]{fontenc}
\usepackage{lmodern}
\usepackage{booktabs}    
\usepackage{multirow}    
\usepackage{array}      
\usepackage{adjustbox}
\usepackage{amssymb}
\usepackage{amsmath}
\usepackage{tikz}
\usepackage{siunitx}
\usepackage{graphicx}
\usepackage[left=2.5cm,right=2.5cm,top=2.5cm,bottom=2.5cm]{geometry}
\usepackage{hyperref}
\usepackage{natbib}
\DeclareMathOperator*{\argmax}{arg\,max}

\begin{document}
% Title and authors
\begin{center}
\rule{\linewidth}{0.5pt}
\vspace{0.5em}
{\Large \textbf{Neural Approaches to SAT Solving: Design Choices and Interpretability}}
\vspace{0.5em}
\rule{\linewidth}{0.5pt}
\vspace{1em}
\textbf{David Mojžíšek$^{2}$, Jan Hůla$^{1,2}$, Ziwei Li$^1$, Ziyu Zhou$^1$, Mikoláš Janota$^1$}\\
\vspace{0.5em}
$^1$Czech Institute of Informatics, Robotics and Cybernetics, Czech Technical University in Prague, Czechia\\
$^2$University of Ostrava, Ostrava, Czechia\\
\vspace{0.5em}
\texttt{david.mojzisek@osu.cz}
\end{center}
% Abstract
\begin{abstract}
In this contribution, we provide a comprehensive evaluation of graph neural networks applied to Boolean satisfiability problems, accompanied by an intuitive explanation of the mechanisms enabling the model to generalize to different instances. We introduce several training improvements, particularly a novel closest assignment supervision method that dynamically adapts to the model's current state, significantly enhancing performance on problems with larger solution spaces. Our experiments demonstrate the suitability of variable-clause graph representations with recurrent neural network updates, which achieve good accuracy on SAT assignment prediction while reducing computational demands. We extend the base graph neural network into a diffusion model that facilitates incremental sampling and can be effectively combined with classical techniques like unit propagation. Through analysis of embedding space patterns and optimization trajectories, we show how these networks implicitly perform a process very similar to continuous relaxations of MaxSAT, offering an interpretable view of their reasoning process. This understanding guides our design choices and explains the ability of recurrent architectures to scale effectively at inference time beyond their training distribution, which we demonstrate with test-time scaling experiments.
\end{abstract}
\section{Introduction}
\label{sec:intro}
Reasoning is a cognitive ability which allows humans to solve problems with previously unseen combinations of constraints. For a long time, it has been debated whether artificial neural networks can obtain such generalization skills or whether they can only learn to detect superficial patterns \cite{fodor1988connectionism,marcus2003algebraic,marcus2018deep} without being able to generalize to novel combinations of constraints. With the arrival of Large Language Models (LLMs) specially trained for reasoning \cite{guo2025deepseek,jaech2024openai}, it became harder and harder to claim that these models can only detect superficial patterns. Nevertheless, the exact mechanism by which they are able to solve tasks that typically require reasoning is largely unknown and the robustness of the solving process is also not understood. 

In this contribution, we focus on a restricted class of problems that require reasoning, concretely on solving Boolean formulas in CNF form. This could be viewed as a prototypical task where the goal is to solve problems with novel combinations of constraints, and where detecting superficial patterns seen during training would be insufficient. It has already been demonstrated that Graph Neural Networks (GNNs) can successfully learn to solve such problems and generalize to larger problems \cite{selsam2018learning}, even though they are still not competitive when compared to state of the art SAT solvers.

Understanding the underlying mechanisms GNNs employ to successfully solve problems, as well as their limitations, would offer significant practical and theoretical value. On the practical side, the trained model can be used as a guessing heuristic inside classical solvers, improving their performance and on the theoretical side, understanding how a GNN can solve a CNF formula could help us to elucidate the reasoning ability of Transformers \cite{vaswani2017attention} because Transformers can be viewed as GNNs in which the graph connectivity is given by the attention map and is learned from data \cite{cai2023connection}. 
Our aim in this contribution is to provide an experimental evaluation of different design choices for GNNs in the context of Boolean satisfiability together with an intuitive explanation of the inner workings of these models. Our main contributions are as follows:

\begin{itemize}
    \item We provide an experimental comparison of different architectures and training regimes.
    \item We introduce a novel supervision method based on the closest assignment, resulting in significant improvements.
    \item We demonstrate that these architectures scale well at test time.
    \item We extend the graph neural network to a diffusion model and show how it relates to the base model.
    \item We provide an intuitive explanation for the inner workings of these models.
\end{itemize}

The rest of the text has the following structure: Section~\ref{sec:bcgr} (Relevant Background) provides the necessary context on Boolean satisfiability problems, SAT solving approaches, graph neural networks, theoretical connection to approximation algorithms, and diffusion models. Section~\ref{sec:exsetup} (Experimental Setup) describes our methodology, including data representation choices, architecture variants, supervision methods, and benchmark generation. Section~\ref{sec:exresult} (Experimental Results) presents our comprehensive evaluation, comparing different graph representations and training methods (Section~\ref{sec:quantitative}), demonstrating test-time scaling capabilities (Section~\ref{sec:testtime}), and introducing our diffusion model extension (Section~\ref{sec:diffext}). Section~\ref{sec:interp} (Interpreting the Trained Model) offers analysis of the embedding space and explains the networks' behavior through the lens of approximation algorithms based on continuous relaxation. Section~\ref{sec:related} (Related Work) positions our contribution within the broader research landscape, and Section ~\ref{sec:discuss} contains a discussion of our findings and directions for future research. We conclude in Section~\ref{sec:concl}. Additional implementation details and mathematical derivations are provided in the Appendix.

\section{Related Work}
\label{sec:related}
Our research builds directly upon NeuroSAT~\cite{selsam2018learning}, which introduced the first end-to-end neural approach for SAT solving using a recurrent message-passing architecture. While we maintain the core iterative design of NeuroSAT (allowing variable numbers of message-passing iterations through weight sharing), we explore simplified variants using RNNs and LSTMs and incorporate techniques like curriculum learning to improve training efficiency.

Several other works have explored different directions in neural SAT solving. Li et al.~\cite{li2023g4satbench} developed G4SATBench to benchmark various GNN architectures (GCN, GGNN, GIN) across different graph representations and supervision objectives. Unlike their broader exploration across architecture types, our work focuses on the recurrent message-passing paradigm from NeuroSAT and investigates how different training objectives and graph representations affect performance within this specific framework. We also mention the work by Warde et al. \cite{warde2023solving} who developed a recurrent architecture based on a Restricted Boltzmann Machine.

Hybrid approaches that integrate neural networks with traditional solvers include NeuroCore by Selsam and Bjørner~\cite{selsam2019guiding}, which uses neural predictions to guide variable branching in CDCL solvers. Similarly, Wang et al.~\cite{wangneurocomb} proposed NeuroComb to enhance CDCL solvers through GNN-based identification of important variables and clauses. These approaches differ from our end-to-end model but demonstrate alternative applications of neural methods to SAT solving.

The connection between neural networks and continuous relaxations is particularly relevant to our work. Kyrillidis et al.~\cite{kyrillidis2020fouriersat} introduced FourierSAT, which transforms Boolean SAT problems into continuous optimization using the Walsh-Fourier transform. This approach provides a theoretical foundation for understanding how neural networks might implicitly convert discrete search problems into continuous optimization. Similar technique was introduced by Hosny et al. ~\cite{hosny2024torchmsat} whow develop GPU-accelerated approaches for MaxSAT problems. Hula et al. \cite{huula2024understanding} and Yau et al.~\cite{yau2024graph} explore the connection between GNNs and semidefinite programming relaxations, demonstrating empirically and theoretically that message-passing can implement gradient-based optimization of SDP relaxations.

In the broader domain of combinatorial optimization, Sun et al.~\cite{sun2023difusco} used diffusion models based on GNNs to solver problems such as traveling salesman. 

\section{Relevant Background}
\label{sec:bcgr}
% ===== SAT/MaxSAT Section =====
\subsection{Boolean Satisfiability and Maximum Satisfiability}
\subsubsection{Boolean Satisfiability as a Constraint Satisfaction Problem}
Boolean satisfiability (SAT) is a fundamental problem in computer science that asks whether a given Boolean formula has a satisfying assignment. The formula is built from propositional variables $x_1, x_2, \dots$ that can take values from $\{0,1\}$, representing false and true respectively, and logical connectives: conjunction ($\land$), disjunction ($\lor$), and negation ($\lnot$). While other connectives like implication ($\rightarrow$) and equivalence ($\leftrightarrow$) exist, they can be expressed using these basic operators.

A literal is either a propositional variable $x$ or its negation $\lnot x$. While Boolean formulas can take arbitrary form, the most common representation is the conjunctive normal form (CNF), where a formula is a conjunction of clauses, and each clause is a disjunction of literals. For example, $(x_1 \lor \lnot x_2) \land (x_2 \lor x_3)$ is a CNF formula with two clauses. We note that any Boolean formula can be transformed into an equisatisfiable CNF formula, albeit potentially requiring additional variables.

An assignment $\sigma$ maps each propositional variable to either 0 or 1. We say $\sigma$ satisfies a CNF formula if at least one literal in each clause evaluates to true under $\sigma$. For instance, the assignment $\sigma(x_1) = 1, \sigma(x_2) = 0, \sigma(x_3) = 1$ satisfies the formula $(x_1 \lor \lnot x_2) \land (x_2 \lor x_3)$ as both clauses contain a true literal.

SAT is a special case of the more general Constraint Satisfaction Problem (CSP) framework \cite{brailsford1999constraint}. A CSP consists of a set of variables, each with a domain of possible values, and a set of constraints that specify allowed combinations of values for groups of variables. While SAT variables are restricted to Boolean values and constraints take the form of clauses, CSPs can accommodate richer variable domains and constraint types.

\subsubsection{MaxSAT: The Optimization Variant}
The Maximum Satisfiability problem (MaxSAT) is the optimization version of SAT . Given a CNF formula $\phi$, the goal is to find an assignment that maximizes the number of satisfied clauses. This formulation is particularly useful when a formula is unsatisfiable, as MaxSAT still yields the best possible solution.

MaxSAT has several variations that differ in their expressiveness and the way they handle the importance of clauses. In unweighted MaxSAT, all clauses have equal importance. Weighted MaxSAT assigns a positive weight to each clause, with the objective being to maximize the sum of weights of satisfied clauses. In some variants (Partial MAX-SAT \cite{fu2006solving}), clauses are categorized as hard or soft, where hard clauses must be satisfied (often with infinite weight), and soft clauses are those that can be violated but contribute to the objective based on their weight. While weighted variants exist, in this paper we focus exclusively on the unweighted formulation.

Formally, for a CNF formula $\phi = c_1 \land c_2 \land \cdots \land c_m$ with $m$ clauses, the unweighted MaxSAT problem seeks an assignment $\sigma^*$ such that:
\begin{equation}
\sigma^* = \argmax_{\sigma} \sum_{i=1}^{m} \mathbf{1}(\sigma \text{ satisfies } c_i)
\end{equation}
where $\mathbf{1}(\cdot)$ is the indicator function.

\subsection{SAT Solving Approaches}
SAT solving algorithms are generally categorized into complete and incomplete approaches, each with distinct characteristics and applications.

\paragraph{Complete Solvers}
Complete solvers theoretically guarantee definitive answers: either finding a satisfying assignment or proving that none exists. The Davis-Putnam-Logemann-Loveland (DPLL) algorithm forms the foundation for most modern complete solvers \cite{biere2009handbook}. It systematically explores the search space through backtracking while employing unit propagation to deduce logical consequences.

Conflict-Driven Clause Learning (CDCL) extends DPLL by analyzing conflicts to learn new clauses, which helps prune large portions of the search space. When a conflict occurs, the solver identifies the "reasons" for the conflict and adds a new clause that prevents similar conflicts in the future. Modern CDCL solvers incorporate sophisticated heuristics for variable selection, restart strategies, and efficient data structures to improve performance.

\paragraph{Incomplete Solvers}
Incomplete solvers focus on finding satisfying assignments but cannot prove unsatisfiability. These algorithms are particularly effective for large satisfiable instances where complete methods might be inefficient.

Local search algorithms, such as WalkSAT \cite{selman1993local}, start with a random assignment and iteratively modify it to satisfy more clauses. These methods employ heuristics to decide which variables to flip at each step, balancing between greedy choices and random moves to escape local optima. For MaxSAT, local search algorithms often use scoring functions that prioritize flipping variables that maximize the increase in satisfied clauses.

Stochastic algorithms including simulated annealing and genetic algorithms have also been applied to SAT and MaxSAT problems. These approaches can effectively explore search spaces in certain problem classes where deterministic methods struggle.

\subsubsection{Continuous Relaxations}
A specific type incomplete solvers that have been explored in recent research \cite{kyrillidis2020fouriersat,hosny2024torchmsat} are explored continuous relaxations of MaxSAT, that transform the discrete problem into a continuous optimization task. These methods map Boolean variables to continuous domains, enabling the application of gradient-based optimization techniques. The Fourier-SAT method \cite{kyrillidis2020fouriersat}, for instance, transforms Boolean formulas into multilinear polynomials through Walsh-Fourier transform and then optimize the continuous variables w.r.t. the resulting polynomial.

Continuous relaxations can also be obtained by making the objective function convex as often done when designing approximation algorithms which provide guarantees for their performance.
The guarantees can be improved by lifting the variables into a high-dimensional vector space and optimizing vectors instead of scalar values. The optimized vectors are finally rounded to discrete values.

Semidefinite Programming (SDP) relaxation, particularly for MAX-2-SAT or MAX-3-SAT, illustrates this approach elegantly. In SDP relaxation, Boolean variables $x_i \in \{0,1\}$ are transformed into unit vectors $\mathbf{y}_i$ in a high-dimensional space. An additional vector $\mathbf{y}_0$ is introduced to represent the value "true." The Boolean variable $x_i$ is considered true if $\mathbf{y}_i$ is close to $\mathbf{y}_0$ (positive inner product) and false if it is far from $\mathbf{y}_0$ (negative inner product).

The optimization process for these vectors follows a pattern:
\begin{enumerate}
    \item Initialize random unit vectors for each variable
    \item Optimize these vectors to maximize the number of satisfied clauses, expressed as a function of inner products between vectors
    \item Round the resulting vectors to discrete assignments (typically based on the sign of inner products with $\mathbf{y}_0$)
\end{enumerate}

This relaxation enables the application of powerful continuous optimization techniques while providing approximation guarantees. For MAX-2-SAT, this approach yields an approximation ratio of 0.878, meaning the solution will satisfy at least 87.8\% of the maximum possible number of clauses.

\subsubsection{Learning-Based Approaches}

Machine learning (ML) is also being heavily utilized for SAT solving. Many approaches have been developed to guide traditional solvers \cite{selsam2019guiding, yolcu2019learning} or to solve SAT problems directly \cite{selsam2018learning, li2022nsnet}. To guide a solver, a neural network can be used to replace heuristics such as variable selection or restart policies. Importantly, Graph Neural Networks (GNNs) can also be trained to solve SAT problems end-to-end without relying on traditional algorithmic solvers \cite{amizadeh2018learning}. These GNN-based approaches can operate directly on the graph representation of Boolean formulas, with variables and clauses forming nodes in a bipartite graph, and learn to predict satisfiability or produce satisfying assignments \cite{li2023g4satbench}. In this work, we focus on variants of GNNs that are recurrent and this allows us to scale the computation during inference or adapt the number of iterations for each instance separately.

In Section \ref{sec:interp} we will show an evidence that one can view the end-to-end ML approaches as bi-level optimization methods because during inference, the GNN behaves as a continuous solver trying to maximize the number of satisfied clauses. Therefore, during training, the outer loop of the bi-level optimization optimizes the weights of the network which then runs an inner loop that optimizes the values of variables to maximize the number of satisfiable clauses.

% ===== GNN Section =====
\subsection{Graph Neural Networks}
Graph Neural Networks (GNNs) extend deep learning to graph-structured data, enabling learning on irregular data structures that classical neural architectures cannot directly process. A graph $G = (V,E)$ consists of nodes $V$ and edges $E$, where each node $v \in V$ may have associated features $x_v$.

GNNs compute node representations through message passing, where each node iteratively aggregates information from its neighbors and updates its features. Formally, at layer $l$, a node $v$ updates its representation $h_v^l$ according to:

$$h_v^{l+1} = \text{UPDATE}(h_v^l, \text{AGGREGATE}(\{h_u^l : u \in \mathcal{N}(v)\}))$$

where $\mathcal{N}(v)$ denotes the neighbors of node $v$. The UPDATE and AGGREGATE functions are typically neural networks, often implementing permutation-invariant operations like sum or max. Through multiple layers of message passing, GNNs can capture both local structure and longer-range dependencies in the graph, making them suitable for processing SAT formulas represented as bipartite graphs.

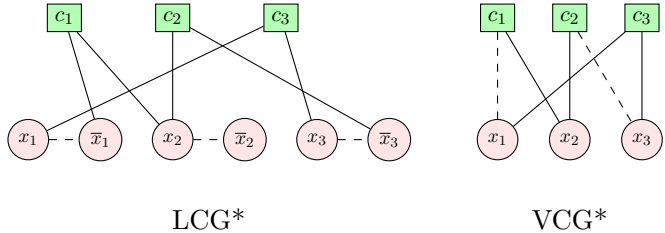
\begin{figure}
    \centering
    \begin{tikzpicture}[scale=0.95, transform shape] % Scaled to fit better
% LCG* (left side)
% Clause nodes
\node[draw, rectangle, fill=green!30, scale=0.8] (c1l) at (-2.5, 2.5) {$c_1$};
\node[draw, rectangle, fill=green!30, scale=0.8] (c2l) at (-1, 2.5) {$c_2$};
\node[draw, rectangle, fill=green!30, scale=0.8] (c3l) at (0.5, 2.5) {$c_3$};

% Literal nodes
\node[draw, circle, fill=pink!40, scale=0.7] (x1l) at (-3, 0.8) {$x_1$};
\node[draw, circle, fill=pink!40, scale=0.7] (nx1l) at (-2, 0.8) {$\overline{x}_1$};

\node[draw, circle, fill=pink!40, scale=0.7] (x2l) at (-1, 0.8) {$x_2$};
\node[draw, circle, fill=pink!40, scale=0.7] (nx2l) at (0, 0.8) {$\overline{x}_2$};

\node[draw, circle, fill=pink!40, scale=0.7] (x3l) at (1, 0.8) {$x_3$};
\node[draw, circle, fill=pink!40, scale=0.7] (nx3l) at (2, 0.8) {$\overline{x}_3$};

% Edges between literals and clauses
\draw[-] (c1l) -- (nx1l);
\draw[-] (c1l) -- (x2l);
\draw[-] (c2l) -- (x2l);
\draw[-] (c2l) -- (nx3l);
\draw[-] (c3l) -- (x1l);
\draw[-] (c3l) -- (x3l);

% Dashed edges between complementary literals
\draw[dashed] (x1l) -- (nx1l);
\draw[dashed] (x2l) -- (nx2l);
\draw[dashed] (x3l) -- (nx3l);

% VCG* (right side) - shifted closer
\node[draw, rectangle, fill=green!30, scale=0.8] (c1r) at (3.5, 2.5) {$c_1$};
\node[draw, rectangle, fill=green!30, scale=0.8] (c2r) at (4.5, 2.5) {$c_2$};
\node[draw, rectangle, fill=green!30, scale=0.8] (c3r) at (5.5, 2.5) {$c_3$};

% Variable nodes
\node[draw, circle, fill=pink!40, scale=0.7] (x1r) at (3.5, 0.8) {$x_1$};
\node[draw, circle, fill=pink!40, scale=0.7] (x2r) at (4.5, 0.8) {$x_2$};
\node[draw, circle, fill=pink!40, scale=0.7] (x3r) at (5.5, 0.8) {$x_3$};

% Solid edges between variables and clauses (positive literals)
\draw[-] (c1r) -- (x2r);
\draw[-] (c2r) -- (x2r);
\draw[-] (c3r) -- (x1r);
\draw[-] (c3r) -- (x3r);

% Dashed edges between variables and clauses (negative literals)
\draw[dashed] (c1r) -- (x1r);
\draw[dashed] (c2r) -- (x3r);

% Labels
\node at (-0.5, -0.3) {LCG*};
\node at (4.5, -0.3) {VCG*};

\end{tikzpicture}  % This loads the TikZ code from the file
    \caption{LCG and VCG of the CNF formula $(\overline{x}_1 \lor x_2) \land (x_2 \lor \overline{x}_3) \land (x_1 \lor x_3)$.}
    \label{fig:graph_representations}
\end{figure}

\subsection{Diffusion-based Assignment Generation}
In Section \ref{sec:diffext} will show how the GNNs we use can be extended to diffusion models which have in recent years emerged as a powerful approach for generative modeling across domains \cite{ho2020denoising}. These models learn to transform a random noise distributions (such as multi-variate Gaussian distribution) to complex distributions behind the given domain (i.e., distribution of images of human faces). For practical applications, diffusion models are typically conditioned on an input so that the generated sample has specific characteristics. In our case, we will condition the model by the bipartite graph of the CNF formula.

\subsubsection{Categorical Diffusion Process}
While continuous diffusion models have gained prominence in image generation and other domains, discrete diffusion processes well-suited for combinatorial optimization problems like MAX-SAT, where the state space is inherently discrete. Our approach presented in Section \ref{sec:diffext} leverages a discrete diffusion process with categorical noise to model the generation of variable assignments. We adapt a concrete form of discrete diffusion first presented by Austin et al. \cite{austin2021structured} and later leveraged for combinatorial optimization with GNNs by Sun et al. \cite{sun2023difusco}. 

On a high level, diffusion models are trained to denoise noisy version of the training samples. These noisy versions are obtained by running a forward diffusion process for several steps and the model is then trained to predict the original sample. For a SAT problem with $n$ variables, we represent each variable assignment as a binary value and the vector of these binary values represent the sample. The diffusion process gradually corrupts this sample until it becomes pure noise.

More concretely, the process that progressively adds noise to the initial assignment $\mathbf{x}_0 \in \{0,1\}^n$ over $T$ timesteps, produces a sequence of increasingly more corrupted assignments $\mathbf{x}_1, \mathbf{x}_2, \ldots, \mathbf{x}_T$. For categorical diffusion, this corruption process is defined by a Markov chain with the following transition matrices:

\begin{equation}
\mathbf{Q}_t = 
\begin{pmatrix}
1-\beta_t & \beta_t \\
\beta_t & 1-\beta_t
\end{pmatrix}
\end{equation}

\noindent where $\beta_t \in (0, 1)$ represents the noise schedule, controlling how quickly the assignments become corrupted. The matrix $\mathbf{Q}_t$ defines the probability of transitioning between states at time $t$, with the property that as $t$ approaches $T$, the distribution of $\mathbf{x}_t$ approaches a uniform distribution over all possible assignments. 

To simplify inference, the cumulative transition matrices $\overline{\mathbf{Q}}_t = \mathbf{Q}_1\mathbf{Q}_2\cdots\mathbf{Q}_t$, which directly gives us $p(\mathbf{x}_t|\mathbf{x}_0)$ are being used. For the Boolean case, this allows us to efficiently sample $\mathbf{x}_t$ given $\mathbf{x}_0$ using:

\begin{equation}
p(\mathbf{x}_t|\mathbf{x}_0) = \text{Cat}(\mathbf{x}_t; \mathbf{p} = \tilde{\mathbf{x}}_0 \overline{\mathbf{Q}}_t)
\end{equation}

\noindent where $\tilde{\mathbf{x}}_0 \in \{0,1\}^{n\times 2}$ is the one-hot encoding of $\mathbf{x}_0$, with each variable represented by a vector $(1,0)$ for value 0 or $(0,1)$ for value 1. The $\text{Cat}$ operation refers to the categorical distribution, which samples $\mathbf{x}_t$ based on the probability vector $\mathbf{p}$.

\subsubsection{Learning the Reverse Process}

The core idea of diffusion models is to learn the reverse process - how to gradually denoise a corrupted sample to recover the original data distribution. In our case, we train a GNN to progressively recover a satisfiable assignment $\mathbf{x}_0$ starting from a random initial assignment.
The trained model is used to sample from a distribution $p(\mathbf{x}_{t-1}|\mathbf{x}_t)$ which can be used to obtain a an assignment $\mathbf{x}_0$ from random assignment $\mathbf{x}_T$ as explained bellow. There are multiple ways of training the neural network used in the diffusion model. One can train it to directly model the distribution $p(\mathbf{x}_{t-1}|\mathbf{x}_t)$. In the method introduced by Austin et al. \cite{austin2021structured}, the network is trained to predict the original uncorrupted input $\mathbf{x}_0$ which is then used to sample from the the posterior $p(\mathbf{x}_{t-1}|\mathbf{x}_t)$ using Bayes' rule. This approach provides stronger learning signals during training, as the target $\mathbf{x}_0$ remains fixed regardless of a timestep and we use it within this work.

\subsubsection{Categorical Posterior Sampling}

As mentioned above, our model is trained to predict $\mathbf{x}_0$ directly and we use this prediction during inference to sample $\mathbf{x}_{t-1}$ given $\mathbf{x}_t$. This is accomplished through categorical posterior sampling, which uses the distribution $p_\theta(\mathbf{x}_0|\mathbf{x}_t, t)$ to compute the posterior $p(\mathbf{x}_{t-1}|\mathbf{x}_t, \mathbf{x}_0)$.

By applying Bayes' rule and the Markov property of the diffusion process, we can derive:

\begin{equation}
p(\mathbf{x}_{t-1}|\mathbf{x}_t) \approx \sum_{\mathbf{x}_0} p(\mathbf{x}_{t-1}|\mathbf{x}_t, \mathbf{x}_0)p_\theta(\mathbf{x}_0|\mathbf{x}_t, t)
\end{equation}

For the categorical case, this is computed using:

\begin{equation}
p(\mathbf{x}_{t-1}|\mathbf{x}_t) \approx \sum_{\mathbf{x}_0} \frac{p(\mathbf{x}_{t-1}|\mathbf{x}_0)p(\mathbf{x}_t|\mathbf{x}_{t-1})}{p(\mathbf{x}_t|\mathbf{x}_0)} p_\theta(\mathbf{x}_0|\mathbf{x}_t, t)
\end{equation}
The diffusion model replaces the distribution $p_\theta(\mathbf{x}_0|\mathbf{x}_t, t)$ with a function approximator (GNN in our case) $f_\theta(\mathbf{x_t},t)$
Therefore, we can train the model using a simple procedure (predicting $\mathbf{x}_0$) and during inference, we can use a sampling process (iteratively sampling $\mathbf{x}_{t-1}$ given $\mathbf{x}_t$), which tries to recover a uncorrupted input in several steps. A useful feature of diffusion models is that the number of sampling steps during inference can be chosen by the user after the model is already trained.

\subsubsection{Inference Schedule}
\label{sec:diff_schdl_bcgr}
During inference, we can accelerate the generation process by using fewer denoising steps than were used during training or use more denoising steps with the hope to increase the quality of outputs. The tuple of time steps used for inference $(T, T -1,\dots,t_0)$ is called a \emph{schedule}. The function approximator in the diffusion model is normally conditioned by the sample at a given time step and also the time step itself ($f_\theta(\mathbf{x_t},t)$) but as we show in Section \ref{sec:diff_assignment_connection}, the time step conditioning is not needed. This means that in our case the schedule is defined only by the number of time steps used.

\section{Experimental Setup}
\label{sec:exsetup}
\subsection{Data Representation and Graph Structure}
Boolean formulas in CNF form can be naturally represented as bipartite graphs where clauses and variables (or literals) form two distinct sets of nodes. In this work, we explore two different graph representations:

\paragraph{Literal-Clause Graph (LCG)}
In the literal-clause graph representation, each literal (both positive and negative polarity of a variable) is represented as a separate node. For a formula with $n$ variables, this results in $2n$ literal nodes. Each literal node is connected to all clause nodes containing that literal. Formally, for a CNF formula $\phi$ with variables $x_1, \ldots, x_n$ and clauses $c_1, \ldots, c_m$, we construct a bipartite graph $G_{LC} = (L \cup C, E)$ where:
\begin{itemize}
    \item $L = \{l_1, \ldots, l_n, \overline{l}_1, \ldots, \overline{l}_n\}$ is the set of literal nodes
    \item $C = \{c_1, \ldots, c_m\}$ is the set of clause nodes
    \item $(l_i, c_j) \in E$ if and only if literal $l_i$ appears in clause $c_j$
\end{itemize}

\paragraph{Variable-Clause Graph (VCG)}
In the variable-clause graph representation, each variable (rather than each literal) is represented as a node. For a formula with $n$ variables, this results in exactly $n$ variable nodes. Each variable node is connected to all clause nodes containing either the positive or negative literal of that variable. To retain information about the polarity of literals, we assign edge features $p_{ij} \in \{-1, 1\}$ to each edge $(x_i, c_j)$, where $p_{ij} = 1$ if the positive literal $x_i$ appears in clause $c_j$, and $p_{ij} = -1$ if the negative literal $\overline{x}_i$ appears in clause $c_j$. Formally, we construct a bipartite graph $G_{VC} = (V \cup C, E, P)$ where:
\begin{itemize}
    \item $V = \{x_1, \ldots, x_n\}$ is the set of variable nodes
    \item $C = \{c_1, \ldots, c_m\}$ is the set of clause nodes
    \item $(x_i, c_j) \in E$ if and only if variable $x_i$ appears in clause $c_j$ (in either polarity)
    \item $P: E \rightarrow \{-1, 1\}$ maps each edge to its corresponding polarity
\end{itemize}

Both graph representations capture the structure of the Boolean formula, but they differ in how they handle variable polarity. The literal-clause graph explicitly represents both polarities as separate nodes, which increases the number of nodes but simplifies the message passing process of the GNN. The variable-clause graph is more compact but requires handling polarity information through edge features. For the GNNs we use, the variable-clause graph representation is more computationally efficient than the literal-clause graph, reducing both memory requirements and processing time. This efficiency comes from having half as many variable nodes (compared to literal nodes) and avoiding an expensive operation during message passing as will be described in Section  \ref{sec:archi}.

In our experiments, we compare both representations together with different message passing operations and different training regimes.

\subsection{Architecture Variants} \label{sec:archi}
Our GNN architecture variants are derived from the NeuroSAT architecture \cite{selsam2018learning} which demonstrated the possibility of using GNNs for SAT solving. The main advantage of this architecture is that it is recurrent and therefore the number of message passing iterations is theoretically not limited. This is not the case for the non-recurrent alternatives with fixed number of layers. We will demonstrate the usefulness of this feature in Section (\ref{sec:testtime}).

\paragraph{Node Embeddings}
Each node in the bi-partite graph of the formula is associated with a $d$-dimensional embedding vector ($d=64$ in most of our experiments as a conclusion from an experiment in \ref{appdim}). We initialize these embeddings randomly from a standard normal distribution. For a formula with $n$ variables and $m$ clauses, we have:
\begin{itemize}
    \item In the literal-clause graph: $2n$ literal embeddings $\mathbf{l}_i \in \mathbb{R}^d$ and $m$ clause embeddings $\mathbf{c}_j \in \mathbb{R}^d$
    \item In the variable-clause graph: $n$ variable embeddings $\mathbf{v}_i \in \mathbb{R}^d$ and $m$ clause embeddings $\mathbf{c}_j \in \mathbb{R}^d$
\end{itemize}

\paragraph{Message Passing Mechanism}
The core of our architecture is a two-phase message passing procedure that alternates between updating clause representations and unknown node representations (literals or variables, depending on the graph type). This process is repeated for a configurable number of iterations $T$.

We primarily use an RNN-based update mechanism, where the node embeddings are the hidden states of the RNN that evolve through message passing iterations. For the variable-clause graph, the message passing at iteration $t$ is defined as:

\begin{align}
\label{eq:clause_up}
\mathbf{h}_c^{(t)} &= \text{RNN}_c\left(\sum_{v \in \mathcal{N}(c)} \mathbf{M}_{vc}(\mathbf{h}_v^{(t-1)}, p_{vc}), \mathbf{h}_c^{(t-1)}\right) \\
\label{eq:var_up}
\mathbf{h}_v^{(t)} &= \text{RNN}_v\left(\sum_{c \in \mathcal{N}(v)} \mathbf{M}_{cv}(\mathbf{h}_c^{(t)}, p_{vc}), \mathbf{h}_v^{(t-1)}\right)
\end{align}

Here, $\mathbf{h}_c^{(t)}$ and $\mathbf{h}_v^{(t)}$ are the hidden states that serve as the actual clause and variable node embeddings for clause nodes and variable nodes respectively. $\mathbf{M}_{vc}$ and $\mathbf{M}_{cv}$ are the message transformation functions that operate on the source node embedding and the edge polarity. For the variable-clause graph, we implement these transformation functions as two MLPs that process positive and negative edges differently:

\begin{align}
\mathbf{M}_{vc}(\mathbf{h}_v, p) = 
\begin{cases}
\text{MLP}_{\text{pos}}(\mathbf{h}_v) & \text{if } p > 0 \\
\text{MLP}_{\text{neg}}(\mathbf{h}_v) & \text{if } p < 0
\end{cases}
\end{align}

For the literal-clause graph, the message passing mechanism also uses operation, called ``Flip'' bellow, that enforces the logical relationship between complementary literals:
\begin{align}
\mathbf{h}_c^{(t)} &= \text{RNN}_c\left(\sum_{l \in \mathcal{N}(c)} \mathbf{h}_l^{(t-1)}, \mathbf{h}_c^{(t-1)}\right) \\
\mathbf{h}_l^{(t)} &= \text{RNN}_l\left(\left[\sum_{c \in \mathcal{N}(l)} \mathbf{h}_c^{(t)}, \text{Flip}(\mathbf{h}_l^{(t-1)})\right], \mathbf{h}_l^{(t-1)}\right)
\end{align}
where $[\cdot, \cdot]$ denotes vector concatenation. The $\text{Flip}(\cdot)$ operation exchanges the embeddings of positive literals with their corresponding negative literals and vice versa. The update function for a given literal embedding can therefore take into account the embedding of the complementary literal.

We note, that the $\text{Flip}(\cdot)$ operation incurs a significant computational cost, particularly for large formulas. In contrast, the variable-clause graph representation eliminates this expensive operation by dedicating only one node for each variable and directly encoding its polarity in edge features. This efficiency makes the variable-clause approach particularly well-suited for larger formulas where computational demands become a critical factor.

Apart from the RNN-based update functions, we also experiment with LSTM-based update functions which have been used in the original NeuroSAT architecture \cite{selsam2018learning}.
The LSTM-based updates follow a similar pattern but maintain an additional cell state alongside the hidden state. In Section \ref{sec:quantitative} we show that different update functions are suitable for different settings.

After each update step, we apply L2 normalization to all node embeddings to stabilize training:
\begin{align}
\mathbf{h}_i^{(t)} = \frac{\mathbf{h}_i^{(t)}}{\|\mathbf{h}_i^{(t)}\|_2}
\label{eq:norm}
\end{align}

\paragraph{Node classification}
After $T$ iterations of message passing, we use the final node embeddings to predict variable assignments. For the variable-clause graph, we apply a linear layer to each variable embedding to produce two logits (representing scores for value true and false): $\mathbf{y}_v = \mathbf{W}\mathbf{h}_v^{(T)} + \mathbf{b}$. The assignment is then determined by applying softmax and taking the argmax: $\hat{a}_v = \argmax_i(\text{softmax}(\mathbf{y}_v)_i)$.

For the literal-clause graph, we focus on the embeddings of positive literals only, as they directly correspond to variables. During training, we use cross-entropy loss between these predicted assignments and the ground truth assignments.

For satisfiability prediction, we can determine whether a formula is satisfiable by checking if the predicted assignment satisfies all clauses. The model is thus trained to find assignments that minimize the number of unsatisfied clauses, effectively solving the MaxSAT problem even when trained only with assignment supervision.

\subsection{Supervision Tasks and Objectives}
There are several obvious supervision objectives and prediction tasks which can be used to train the model. The original NeuroSAT model was trained to predict the satisfiability status of a given formula using binary cross-entropy. Later, several authors tried different training tasks and objectives which have been summarized in a review paper by Li et al. \cite{li2023g4satbench}. We reimplement these objective and task for our setup and also introduce a novel training objective which in certain settings results in significant improvements of the model performance. These objective are briefly described below.

\paragraph{Satisfiability Classification}
This is the task which was used by \cite{selsam2018learning} for training the original NeuroSAT architecture. The model is trained to predict whether the formula is satisfiable or not through graph-level embedding aggregation using global mean pooling. The loss is computed by binary cross-entropy between the prediction $\hat{y}$ and ground truth $y \in \{0, 1\}$: $
\mathcal{L}_{\text{sat}} = -(y\log\hat{y} + (1-y)\log(1-\hat{y}))
$.

\paragraph{Unsupervised Training}
For unsupervised training, we define the loss using clause validity \cite{ozolins2022goal}, where $\hat{x}_i$ represents the model's predicted continuous probability of a variable being true:
\begin{align}
V_c(\hat{x}) = 1 - \prod_{i\in c^+}(1 - \hat{x}_i)\prod_{i\in c^-}\hat{x}_i, \quad \mathcal{L}_{\phi}(\hat{x}) = -\sum_{c\in\phi}\log(V_c(\hat{x})),
\end{align}
where $c^+$ and $c^-$ are the sets of variables that occur in clause $c$ in positive and negative form respectively. This loss reaches its minimum only when the prediction $\hat{x}$ is a satisfying assignment. We note that alternative unsupervised formulations exist \cite{amizadeh2018learning}, and comprehensive evaluations reported by Li et al. \cite{li2023g4satbench} suggest that these two different approaches perform similarly in practice. Another training option would be to directly optimize a convex loss function derived from SDP relaxation, but this approach is limited because SDP formulations work well for MAX-2-SAT and can be extended to MAX-3-SAT, but become increasingly difficult to formulate for general MaxSAT problems with larger clauses.

\paragraph{Assignment Prediction}
For satisfiable formulas, we can train the model to predict the satisfiable variable assignments directly. We tried to use either mean squared error or cross-entropy loss between the predicted assignments and the ground truth assignments: $
\mathcal{L}_{\text{assign}}^{MSE} = \|\hat{a} - x\|_2^2 \quad \text{and} \quad \mathcal{L}_{\text{assign}}^{CE} = -\sum_i x_i\log\hat{x}_i + (1-x_i)\log(1-\hat{x}_i)
$
where $x$ is the ground truth assignment and $\hat{a},\hat{x}$ are the predicted assignments which differ by application of softmax (i.e. $\hat{a}$ are just logits without a softmax applied).

\paragraph{Closest Assignment Training}
One problem with assignment prediction is that satisfiable formulas can have a lot of solution and the network is penalized even if it predicts satisfiable solution which differs from the one which is used as a ground truth. We therefore introduce a novel supervision method which uses a MaxSAT solver to always compute the solution which is closest to the solution predicted by the model. We then update then model with respect to this solution. In Section \ref{sec:quantitative}, we show that this method works particularly well when the solution space is large.

For each formula in a batch, a valid assignments that minimize the Hamming distance to the model's current predictions is found by the RC2 MaxSAT solver. For satisfiable formulas it finds an assignment that satisfies all clauses while being closest to current prediction. For unsatisfiable formulas, it finds an assignment that maximizes the number of satisfied clauses while minimizing distance to prediction.

This approach allows the model to explore different regions of the solution space while maintaining valid solutions for SAT instances or optimal partial solutions for UNSAT instances. The supervision signal adapts to the model's current state rather than forcing it toward a single pre-determined assignment. The disadvantage of this method is that the computation of the loss is slower then with the precomputed solution. This could be solved by pre-computing solutions or by using an approximate MaxSAT solver.

\paragraph{SAT-Only Instance Filtering}
After initially training with both satisfiable and unsatisfiable instances, we experimented with formula-type specialization by restricting training to only satisfiable instances. In Table \ref{tab:vcg_dataset_comparison}, we show that this filtering can lead to higher accuracy of the trained model.

\subsection{Benchmarks and Data Generation}\label{datadecs}

We utilize two complementary benchmark generators for evaluating the tested variants: the SR generator and a 3-SAT generator with the ratio between variable and clauses set close to the phase transition point.

\paragraph{SR Generator}
The SR generator by Selsam et al. \citep{selsam2018learning} produces pairs of satisfiable and unsatisfiable formulas that differ by negating only a single literal. This design specifically prevents models from exploiting superficial features for classification. Intuitively, it works by iteratively sampling random clauses and adding them to a formula. After each addition, a SAT solver checks if the formula remains satisfiable. When adding a clause that finally makes the formula unsatisfiable, the generator saves this instance and creates its satisfiable counterpart by flipping a single literal in the last clause. To create each clause, it samples a small integer $k$ based on a mix of Bernoulli and geometric distributions, then randomly selects $k$ variables without replacement, negating each with 0.5 probability. This solver-driven approach ensures that satisfiability classification requires understanding the logical structure rather than statistical properties. As reported in the review by Li et al. \cite{li2023g4satbench}, the models trained on problems from this generator transfer the best to other problem distributions. 

\paragraph{3-SAT Generator}
We also employ a 3-SAT generator configured at the critical clause-to-variable ratio of 4.26, known as the phase transition point where SAT problems are empirically the most challenging to solve \citep{crawford1996experimental}. At this ratio, approximately half of the generated instances are satisfiable. Each clause contains exactly 3 literals selected uniformly from the available variables, with each literal negated with 0.5 probability. Unlike the SR generator, 3-SAT focuses on generating naturally difficult problems rather than explicitly preventing superficial feature learning.

\section{Experimental Results}
\label{sec:exresult}

\subsection{Training and Evaluation Methodology}
For training, we generate 50,000 instances: 25,000 pairs for SR and 50,000 instances for 3-SAT. We annotate each dataset by the maximum number of variables appearing in the training formulas. For SR, we test two variations, SR40 for which the training examples are sampled with 3-40 variables and SR100 for which the training examples contain 10-100 variables. For 3-SAT, the training samples contain 10-100 variables (3SAT100). The SR dataset is well suited for training SAT/UNSAT prediction models due to its design that prevents learning from superficial features, making it harder for models to exploit statistical shortcuts rather than learning true logical reasoning. We also create versions of training data which contain only satisfiable instances (denoted SAT only). The size of these datasets is half of the original datasets (i.e. 25000 examples). To evaluate generalization, we validate exclusively on problems with exactly the maximum number of variables in each category, therefore SR40 for evaluation means that the problems have always exactly 40 variables (not a range of 3-40), SR100 test contains only problems with exactly 100 variables, and so on.\footnote{This is consistent with the original experiments by Selsam et al. but different from the experiments reported by Li et al. \cite{li2023g4satbench} where SR40 in evaluation contains problems from the same distribution as the training set and therefore they report a better performance because smaller problems are easier to solve.}

Table \ref{tab:benchmark-stats} summarizes the key statistics of our evaluation datasets.

\begin{table}[ht]
\centering
\caption{Statistics of benchmark test sets. SAT\% indicates the percentage of satisfiable instances in each dataset. Avg. Gap represents the average number of unsatisfied clauses when using random variable assignments. SAT Gap and UNSAT Gap show this metric separated by instance satisfiability. SR datasets are generated using the SR generator with the indicated number of variables (e.g., SR40 contains instances with 40 variables), while 3SAT datasets contain instances near the phase transition point with the specified number of variables. All datasets maintain a balanced distribution of satisfiable and unsatisfiable instances.}
\label{tab:benchmark-stats}
\begin{tabular}{@{}lccccc@{}}
\toprule
\textbf{Dataset} & \textbf{SAT\%} & \textbf{Avg. Gap} & \textbf{SAT Gap} & \textbf{UNSAT Gap} & \textbf{Avg. Clauses} \\
\midrule
SR40 & 50.0\% & 21.29 & 21.59 & 20.99 & 228.40 \\
SR100 & 50.0\% & 51.31 & 50.64 & 51.98 & 547.49 \\
SR200 & 50.0\% & 100.31 & 101.03 & 99.59 & 1083.81 \\
SR400 & 50.0\% & 198.74 & 198.53 & 198.95 & 2152.32 \\
\midrule
3SAT100 & 53.5\% & 52.78 & 53.00 & 52.54 & 426.00 \\
3SAT200 & 55.5\% & 107.65 & 107.45 & 107.90 & 852.00 \\
\bottomrule

\end{tabular}
\end{table}

The Gap metric represents the average number of unsatisfied clauses when using random variable assignments. This metric has the same definition for both SAT and UNSAT instances; it simply counts how many clauses remain unsatisfied with random assignments on average. Larger gaps indicate more challenging problems where random guessing performs poorly.

\subsection{Quantitative Evaluation}\label{sec:quantitative}

We conducted a comprehensive evaluation that compares different architectural choices and supervision methods. Our evaluation focuses on five key performance metrics:

\begin{itemize}
    \item \textbf{Average Gap}: The average number of unsatisfied clauses across all test instances. Lower values indicate better performance, with 0 representing perfect satisfaction (i.e., no unsatisfied clauses) on satisfiable instances. For unsatisfiable instances, this metric reflects how close the model gets to minimizing unsatisfied clauses.
    
    \item \textbf{Gap on SAT}: The average number of unsatisfied clauses computed only over satisfiable instances.
    
    \item \textbf{Gap on UNSAT}: The average number of unsatisfied clauses computed only over unsatisfiable instances.
    
    \item \textbf{SAT Accuracy}: The percentage of satisfiable instances for which the model correctly finds a satisfying assignment, computed only over satisfiable instances.
    
    \item \textbf{Decision Accuracy}: The percentage of instances for which the model correctly predicts whether the formula is satisfiable. Since our approach does not formally refute unsatisfiable instances, we classify an instance as unsatisfiable when the model fails to find a satisfying assignment. This means unsatisfiable instances are always classified correctly under this assumption. This applies specifically in the case of assignment-based evaluation.
\end{itemize}

\subsubsection{Comparison of Graph Representations, Update Functions and Training Methods}

Table~\ref{tab:gnn_sat_performance} presents a comprehensive comparison of different architectural configurations trained exclusively on the SR40 dataset. This comparison includes different graph representations (Literal-Clause Graph vs. Variable-Clause Graph), update functions (RNN vs. LSTM), and supervision approaches (SAT/UNSAT classification, assignment supervision, and unsupervised objective training), all evaluated on instances with 40 variables. All models were evaluated using Exponential Moving Average (EMA) of parameters during validation only, as detailed in~\ref{aema}, which helps reduce fluctuations in validation metrics and provide more reliable model selection. Importantly, curriculum learning (~\ref{curric}) was employed only for training models with SAT/UNSAT classification objectives, as it proved unnecessary for models trained with assignment prediction or unsupervised learning approaches.

\paragraph{Graph Representation Impact:} Our results demonstrate that Literal-Clause Graph (LCG) and Variable-Clause Graph (VCG) representations exhibit different strengths. VCG shows better performance for assignment-based training with RNN updates, achieving a SAT accuracy of 68.8\% compared to 48.6\% for LCG. Additionally, VCG's more compact representation (using one node per variable rather than two for positive and negative literals) provides computational advantages for larger formulas, making it our preferred choice for scaling to more complex problems.

\paragraph{Message Passing Mechanism:} While LSTM-based message passing shows advantages in some configurations, particularly for unsupervised training, we found that RNN-based approaches offer a better balance of performance and interpretability for assignment-based training. RNN updates with VCG representation achieved higher results for finding satisfying assignments, with 68.8\% SAT accuracy and 84.4\% decision accuracy. The simpler RNN structure also facilitates better analysis of the model's internal reasoning process. However, we found training RNN-based models for SAT/UNSAT classification particularly challenging, with LSTM being more stable for this specific task.

\paragraph{Supervision Approach:} Our experiments reveal distinct advantages for different supervision approaches:
\begin{enumerate}
    \item \textbf{Assignment-based supervision} shows better performance for finding satisfying assignments, especially with VCG+RNN configuration (68.8\% SAT accuracy, 84.4\% decision accuracy).
    
    \item \textbf{Unsupervised learning} achieves the lowest average gaps across configurations (as low as 0.91 for VCG+RNN and 0.84 for VCG+LSTM). This makes unsupervised training useful for applications where minimizing unsatisfied clauses is the priority.
    
    \item \textbf{SAT/UNSAT classification} training, while challenging with RNN, enables an interesting property: models trained only for classification develop an implicit ability to separate embeddings for positive and negative literals. This separation allows for retrieving satisfying assignments through clustering techniques, despite the model not being explicitly trained for assignment prediction.
\end{enumerate}

Based on the results reported in Table \ref{tab:gnn_sat_performance}, we identify the VCG+RNN+Assignment configuration as our most effective approach, offering a good balance between assignment accuracy and computational efficiency. This configuration forms the foundation for our further experiments and analysis in subsequent sections.

\paragraph{Assignment Training Refinements:} Table \ref{tab:vcg_dataset_comparison} highlights the impact of a novel training method we introduce, here called ``closest assignment'', with the VCG+RNN configuration across multiple datasets. This method computes assignments that minimize Hamming distance to the model's current predictions, showing improvements over training with precalculated assignments, especially for formulas with more variables. For SR100, using the closest assignment approach reduces the average gap from 3.81 to 1.43 for SAT+UNSAT training and improves SAT accuracy from 44.8\% to 53.2\%.

This improvement correlates with the number of possible solutions in the benchmarks (SR10-100 has a median of 16 solutions per formula compared to SR3-40's median of 7), supporting our hypothesis that for formulas with larger solution spaces, guiding the model with dynamically selected assignments that align with its current predictions yields better generalization than using fixed predetermined assignments.

The computational challenges of calculating closest assignments during training are noteworthy, particularly for larger benchmarks like 3SAT+UNSAT, where this approach became impractical and we therefore omit this experiment and leave the last row of Table \ref{tab:vcg_dataset_comparison} empty. It also highlights an opportunity for future work on more efficient approximation methods for finding near-optimal assignments.

\paragraph{Training Data Composition:} Our results also indicate that training exclusively on SAT instances (SAT only) improves performance for finding satisfying assignments. For SR40, this approach with closest assignment training achieves our highest SAT accuracy of 76\% and decision accuracy of 88\%. However, models trained on both SAT and UNSAT instances (SAT+UNSAT) with closest assignment supervision demonstrate better gap minimization, achieving an average gap of 0.98 versus 2.68 for SAT-only training on SR40.

\begin{table}[ht]
\centering
\caption{Performance comparison of GNN architectures for SAT solving on the SR40 dataset. The table compares Literal-Clause Graph (LCG) and Variable-Clause Graph (VCG) representations, RNN and LSTM update mechanisms, and different training objectives. Metrics include average gap (number of unsatisfied clauses) across all instances and separated by satisfiability status (lower is better), SAT accuracy (percentage of satisfiable instances solved by finding assignment), and decision accuracy (percentage of correct satisfiability predictions). Notable findings include: unsupervised training consistently achieves lowest gaps; VCG+RNN with assignment prediction shows highest SAT accuracy (68.8\%); and RNN-based models with SAT/UNSAT classification proved challenging to train effectively (indicated by dashes). Asterisks (*) indicate results obtained through clustering of node embeddings rather than direct prediction. This model combination was particularly hard to train in our setup. We found that both for VCG and LCG RNN is very sensitive to hyper-parameter selection. As the model failed to get generalized in our final unified experimental setup we do not include this result (close to random performance) now.}
\label{tab:gnn_sat_performance}
\resizebox{\textwidth}{!}{%
\begin{tabular}{@{}lllcccccc@{}}
\toprule
Graph & Update & Loss Function & Avg. Gap $\downarrow$ & Gap on SAT $\downarrow$ & Gap on UNSAT $\downarrow$ & SAT Acc. $\uparrow$ & Dec. Acc. $\uparrow$ \\
\midrule
\multirow{6}{*}{LCG} & \multirow{3}{*}{RNN} & SAT/UNSAT & --- & --- & --- & --- & --- \\
 & & Assignment & 1.83 & 1.25 & 2.41 & 48.6 \% & 72.8 \% \\
 & & Unsup & 0.93 & 0.59 & 1.26 & 51.4 \% & 75.7 \% \\
\cmidrule{2-8}
 & \multirow{3}{*}{LSTM} & SAT/UNSAT & 1.96* & 1.27* & 2.62* & 59.2* \% & 83.9 / 79.6* \% \\
 & & Assignment & 1.82 & 1.06 & 2.58 & 56.8 \% & 78.4 \% \\
 & & Unsup & \textbf{0.81} & \textbf{0.45} & \textbf{1.16} & 62 \% & 81 \% \\
\midrule
\multirow{6}{*}{VCG} & \multirow{3}{*}{RNN} & SAT/UNSAT & 3.62* & 1.9* & 5.34* & 56.6* \% & 80 / 78.3* \% \\
 & & Assignment & 1.95 & 0.8 & 3.05 & \textbf{68.8} \% & \textbf{84.4} \% \\
 & & Unsup & 0.91 & 0.58 & 1.23 & 51.6 \% & 75.8 \% \\
\cmidrule{2-8}
 & \multirow{3}{*}{LSTM} & SAT/UNSAT & 2.33* & 1.57 & 3.08 & 52.2* \% & 81.9 / 76.1* \% \\
 & & Assignment & 2.05 & 0.96 & 3.14 & 66.4 \% & 83.2 \% \\
 & & Unsup & 0.84 & 0.51 & 1.17 & 56.4 \% & 78.2 \% \\
\bottomrule
\end{tabular}%
}
\end{table}

\begin{table}[ht]
\centering
\caption{Performance analysis of VCG+RNN with assignment prediction across different datasets and training methodologies. Our novel "Closest" supervision method (which dynamically selects assignments closest to current model predictions) consistently outperforms training with precalculated assignments. For SR40, SAT-only training with closest assignment supervision achieves the highest SAT accuracy (76\%), while SAT+UNSAT training with closest assignment supervision yields the lowest average gap (0.98). The missing data for 3SAT100 with SAT+UNSAT closest supervision is due to prohibitive computational costs. Bold values indicate best results per dataset.}
\label{tab:vcg_dataset_comparison}
\resizebox{\textwidth}{!}{%
\begin{tabular}{@{}llcccccc@{}}
\toprule
Dataset & Training Mode & Assignment Type & Avg. Gap $\downarrow$ & Gap on SAT $\downarrow$ & Gap on UNSAT $\downarrow$ & SAT Acc. $\uparrow$ & Dec. Acc. $\uparrow$ \\
\midrule
\multirow{4}{*}{SR40} 
 & SAT only & Precalculated & 2.93 & 1.11 & 4.75 & 68.2 \% & 84.1 \% \\
 & SAT only & Closest & 2.68 & 0.88 & 4.48 & \textbf{76 \%} & \textbf{88 \%} \\
 & SAT+UNSAT & Precalculated & 1.95 & 0.8 & 3.05 & 68.8 \% & 84.4 \% \\
 & SAT+UNSAT & Closest & \textbf{0.98} & \textbf{0.48} & \textbf{1.49} & 71.2 \% & 85.6 \% \\
\midrule
\multirow{4}{*}{SR100} 
 & SAT only & Precalculated & 4.42 & 2.36 & 6.48 & 47.4 \% & 73.7 \% \\
 & SAT only & Closest & 3.57 & 1.67 & 5.48 & \textbf{59.6} \% & \textbf{79.8} \% \\
 & SAT+UNSAT & Precalculated & 3.81 & 2.34 & 5.28 & 44.8 \% & 72.4 \% \\
 & SAT+UNSAT & Closest & \textbf{1.43} & \textbf{0.92} & \textbf{1.94} & 53.2 \% & 76.6 \% \\
\midrule
\multirow{4}{*}{3SAT100} 
 & SAT only & Precalculated & 5.93 & 3.40 & 9.27 & 25.7 \% & 57.2 \% \\
 & SAT only & Closest & 5.23 & \textbf{2.33} & 9.11 & 48,4 \% & \textbf{70} \% \\
 & SAT+UNSAT & Precalculated & \textbf{4.22} & 2.84 & \textbf{6.00} & 23,9 \% & 55.8 \% \\
 & SAT+UNSAT & Closest & --- & --- & --- & --- & --- \\
\bottomrule
\end{tabular}%
}
\end{table}

\begin{figure}[ht]
    \centering
    \begin{minipage}{0.49\linewidth}
        \centering
        \includegraphics[width=\linewidth]{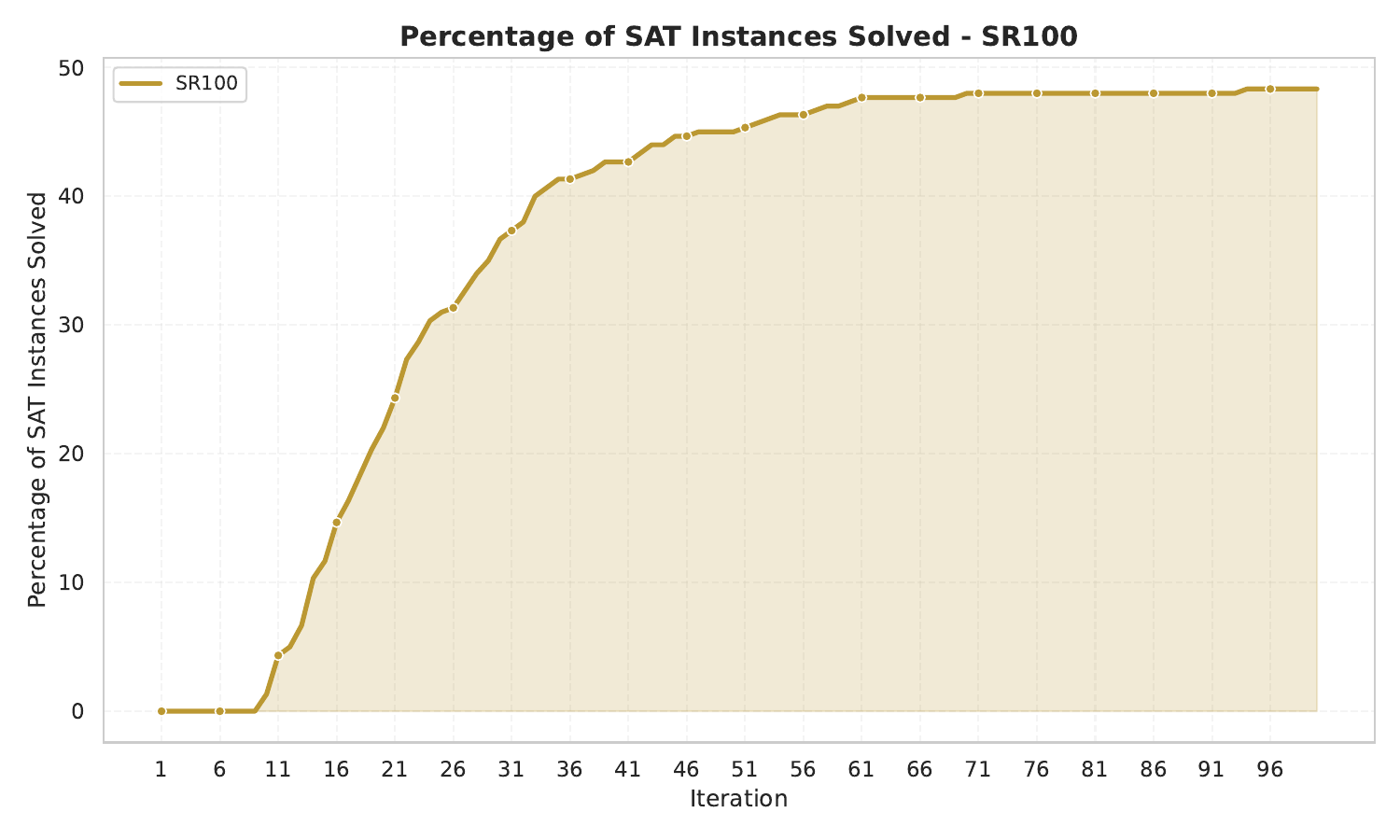}
    \end{minipage}
    \hfill
    \begin{minipage}{0.49\linewidth}
        \centering
        \includegraphics[width=\linewidth]{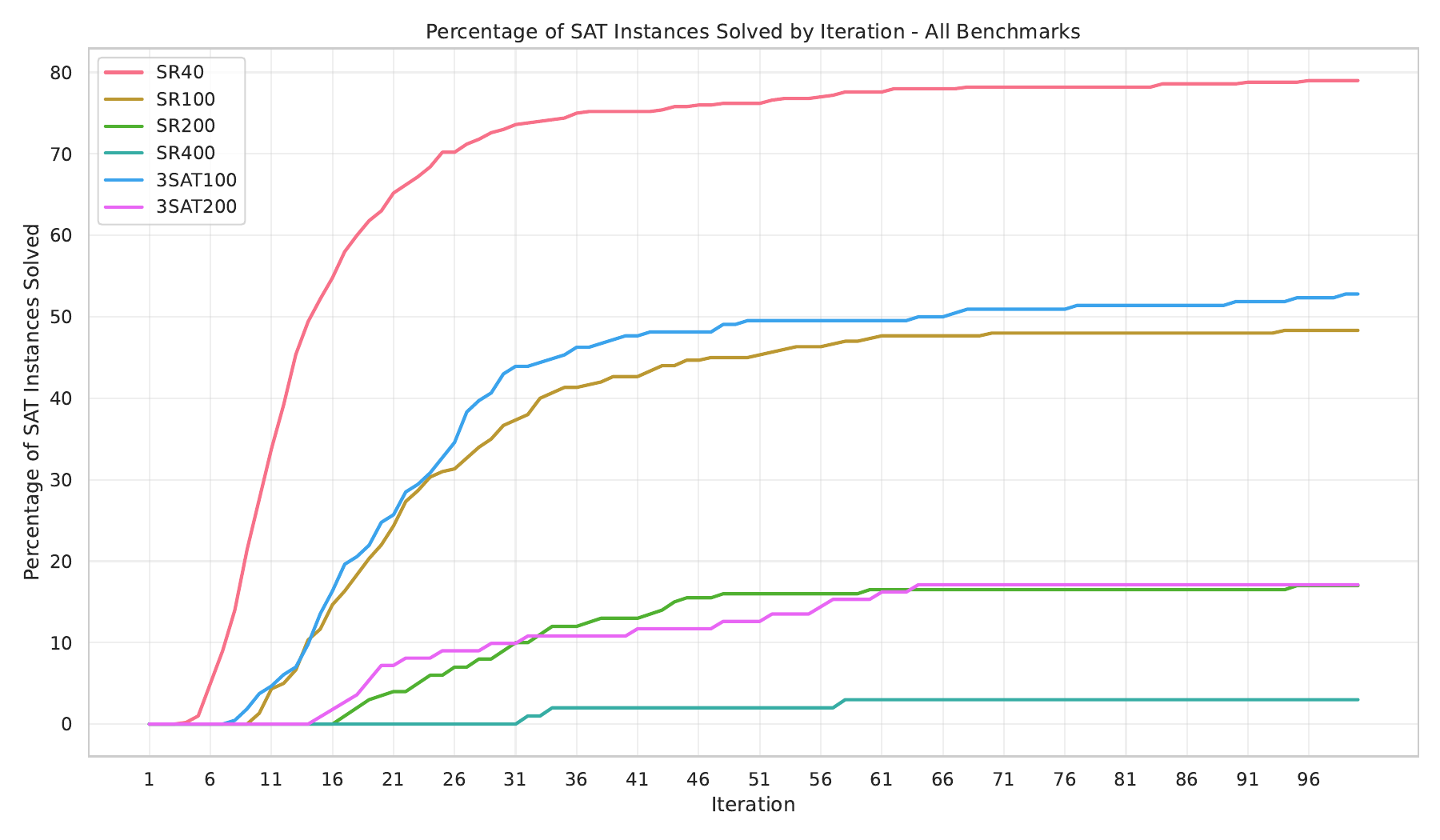}
    \end{minipage}
    \caption{Percentage of SAT instances solved as message passing iterations increase for a model trained on SR40 with SAT+UNSAT closest assignment supervision. Left: Performance on SR100, showing rapid initial improvement. Right: Comparison across benchmarks, demonstrating effectiveness decreases with problem size but benefits from additional iterations, highlighting the recurrent architecture's inference-time scaling capability.}
    \label{fig:sat_evolution}
\end{figure}

\begin{figure}[ht]
    \centering
    \begin{minipage}{0.49\linewidth}
        \centering
        \includegraphics[width=\linewidth]{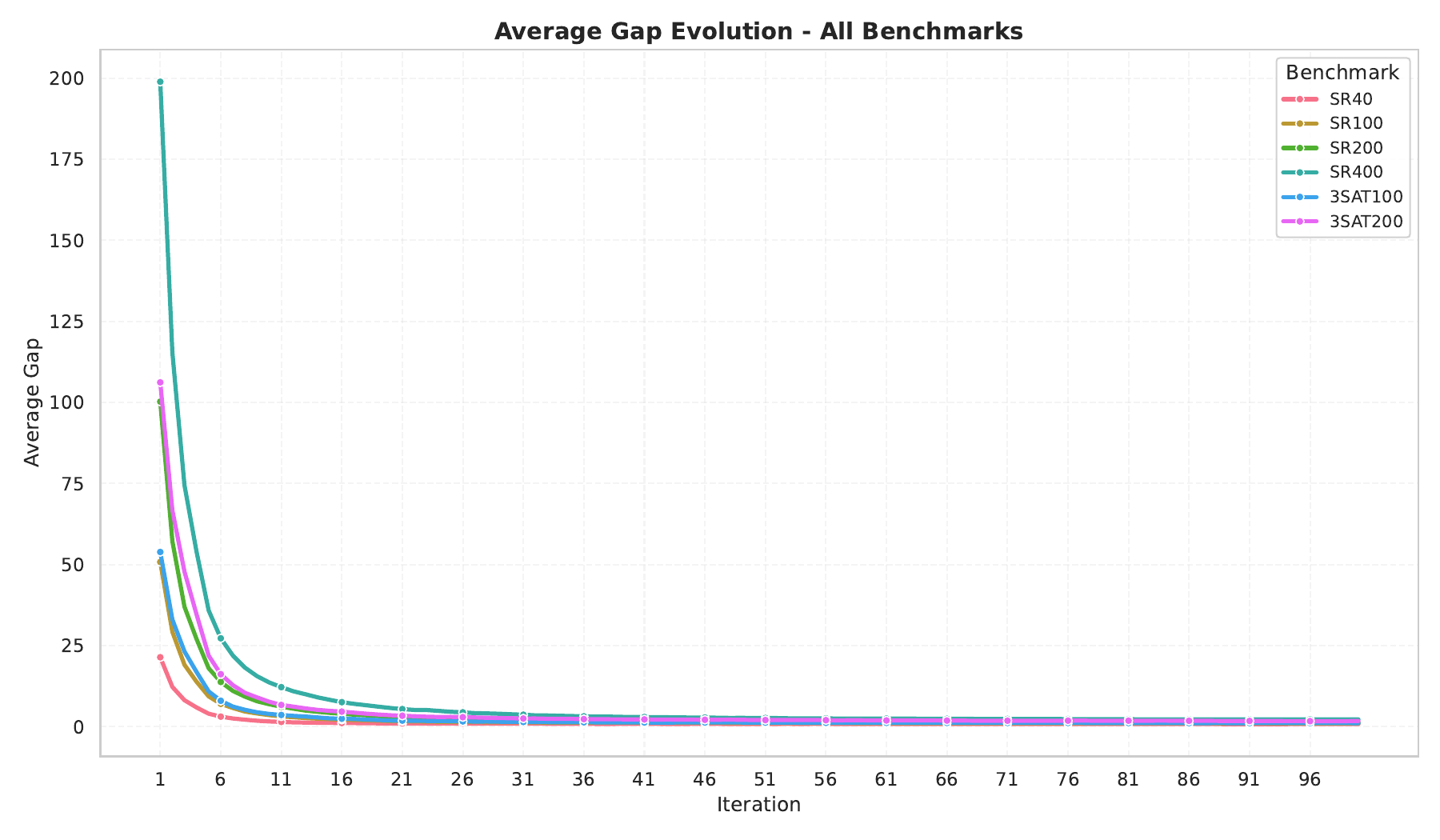}
    \end{minipage}
    \hfill
    \begin{minipage}{0.49\linewidth}
        \centering
        \includegraphics[width=\linewidth]{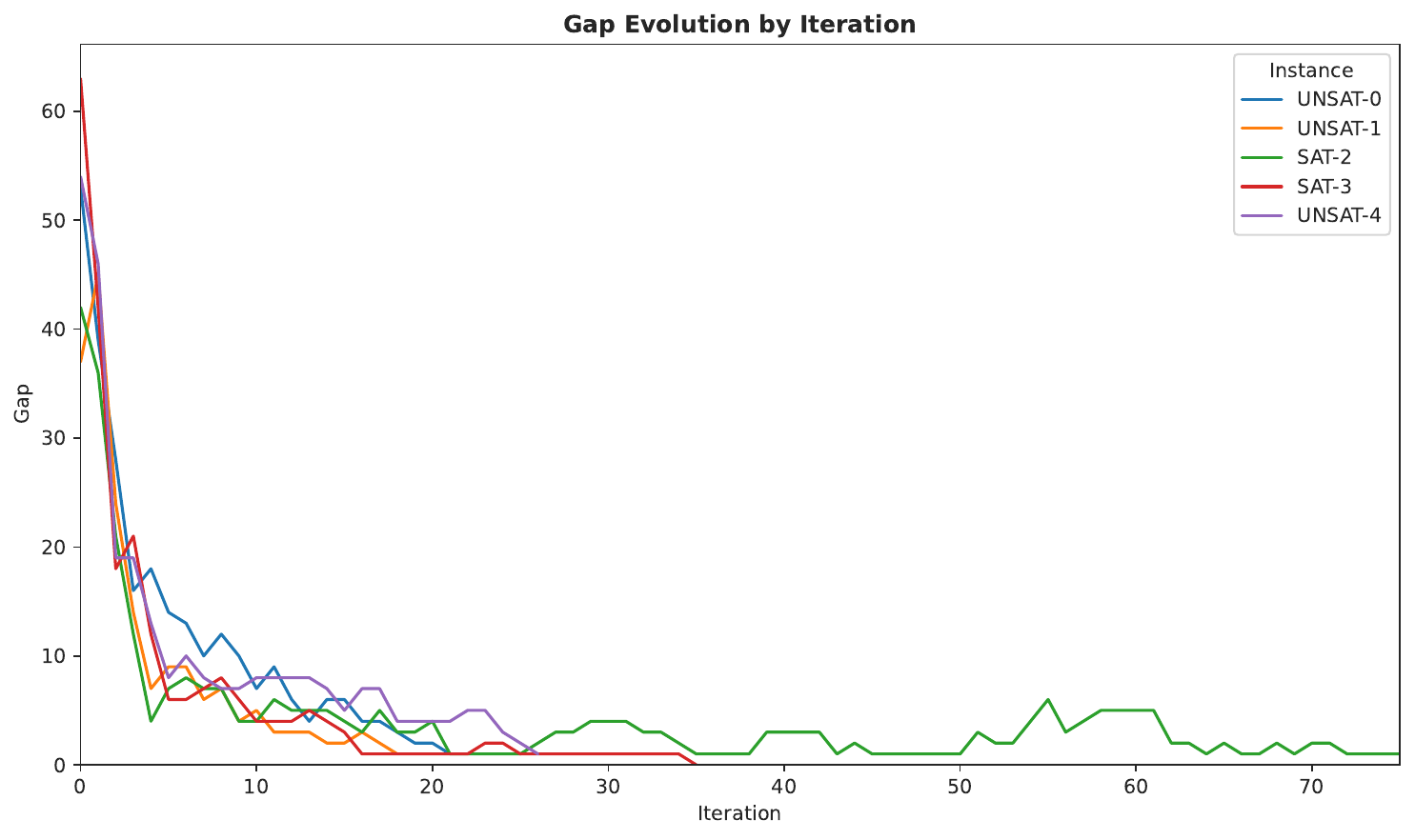}
    \end{minipage}
    \caption{Average gap (unsatisfied clauses) reduction with increasing message passing iterations for a model trained on SR40 with SAT+UNSAT closest assignment supervision. Left: Comparison across benchmarks showing extremely rapid gap reduction in early iterations for all problem sizes, with all benchmarks achieving remarkably low average gaps despite varying SAT-solving performance. Right: Individual instance trajectories revealing different convergence patterns between SAT and UNSAT instances, with occasional fluctuations suggesting potential benefit from monitoring solution quality during inference.}
    \label{fig:gap_evolution}
\end{figure}

\subsection{Test-time Scaling} \label{sec:testtime}

\begin{figure}[ht]
    \centering
    \includegraphics[width=\linewidth]{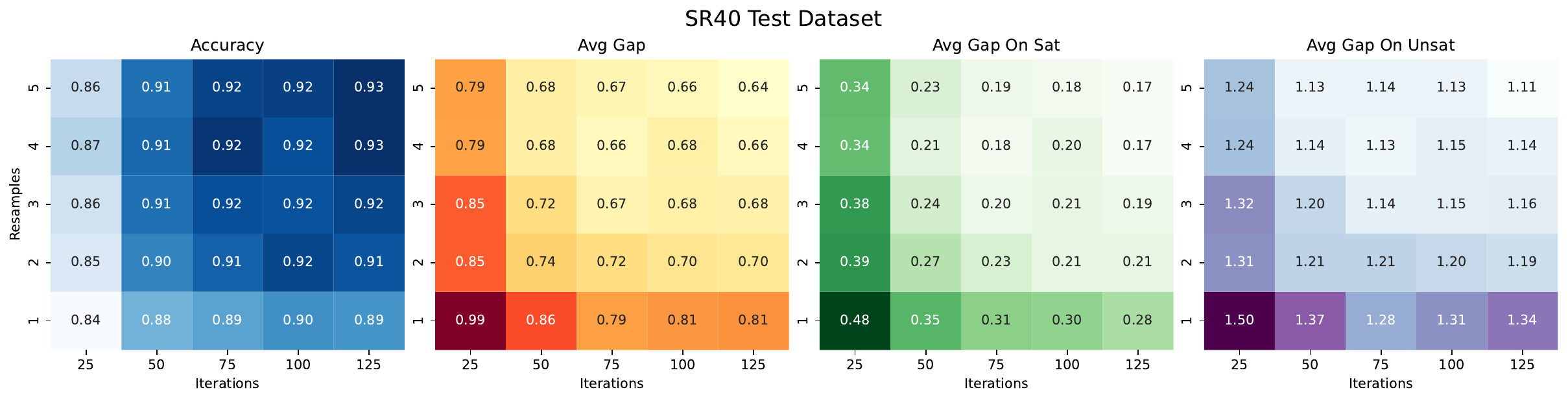}
    \includegraphics[width=\linewidth]{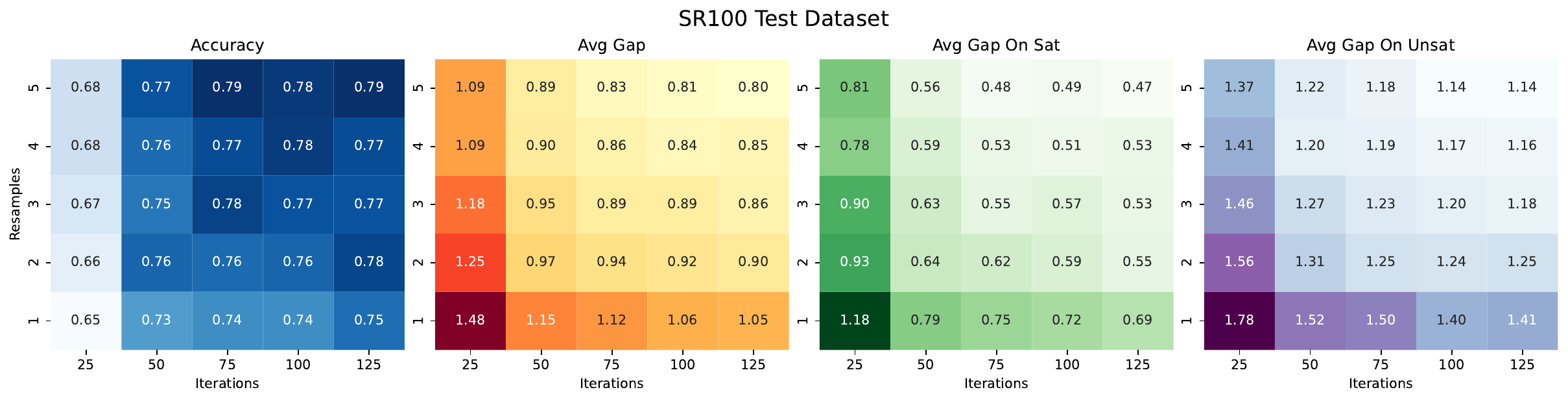}
    \includegraphics[width=\linewidth]{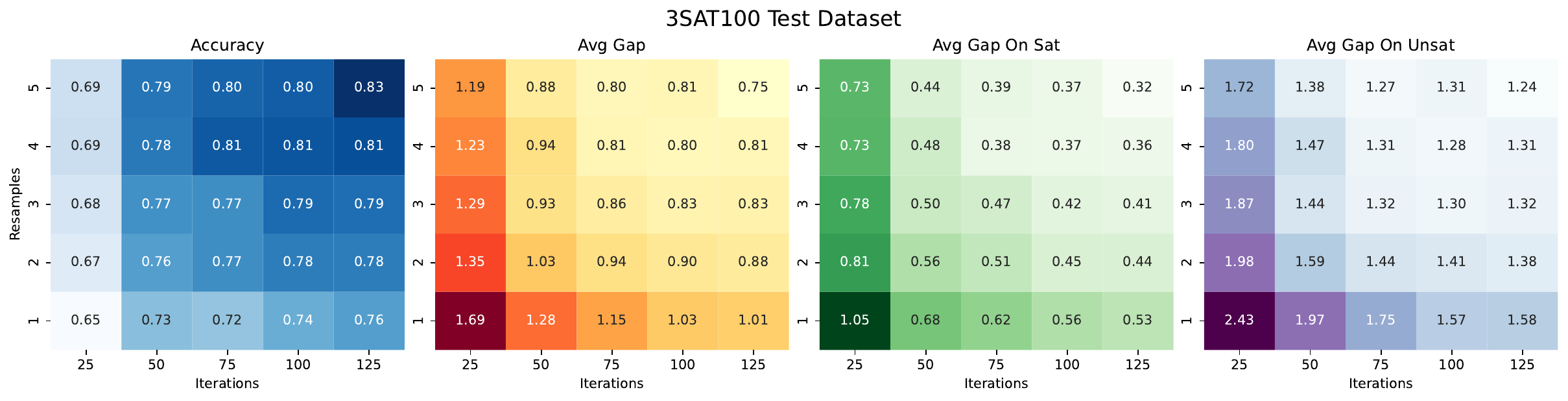}
    \caption{Performance heatmaps for a model trained on SR40 with SAT+UNSAT closest assignment supervision, showing how metrics improve with both increased iterations (columns) and resampling attempts (rows). Testing on SR40 (top), SR100 (middle), and 3SAT100 (bottom) demonstrates significant gains from both scaling dimensions—e.g., SR40 decision accuracy improves from 84\% (1 sample, 25 iterations) to 93\% (5 samples, 125 iterations). This two-dimensional inference-time scaling capability is consistent across benchmarks but with decreasing returns on larger problems.}
    \label{fig:metrics_heatmaps}
\end{figure}

\begin{table}
\centering
\caption{Performance of a model trained on SR40 (VCG+RNN with closest assignment supervision) when tested across various benchmarks with a maximum of 100 message-passing iterations and early stopping. The model maintains reasonable performance on SR100 (74.2\% decision accuracy) but degrades on larger instances. "UNSAT Instances (gap == 1)" shows the percentage of UNSAT instances where the model achieved a gap of 1, which is always optimal for SR datasets but not always achievable for 3SAT instances. "Steps" columns indicate average/median iterations required to reach solutions, demonstrating the model's efficiency.}
\label{tab:mod40onbench}
\begin{tabular}{lccccc}
\hline
Dataset & Decision & SAT Instances & UNSAT Instances & SAT Steps & UNSAT Steps \\
 & Accuracy & Solved & (gap $==$ 1) & (Avg/Med) & (Avg/Med) \\
\hline
SR40 & 89.5\% & 79.0\% & 95.6\% & 16.17/13.0 & 13.33/10.0 \\
SR100 & 74.2\% & 48.3\% & 91.3\% & 24.93/21.0 & 23.47/19.0 \\
SR200 & 58.5\% & 17.0\% & 64.5\% & 32.44/30.0 & 33.98/28.0 \\
SR400 & 51.5\% & 3.0\% & 16.0\% & 41.33/34.0 & 52.06/44.5 \\
3SAT100 & 74.8\% & 52.8\% & 64.0\% & 25.63/22.0 & 29.66/24.0 \\
3SAT200 & 54.0\% & 17.1\% & 22.5\% & 33.21/25.0 & 35.10/31.5 \\
\hline
\end{tabular}
\end{table}

\begin{table}
\centering
\caption{Performance of a model trained on SR100 (VCG+RNN with closest assignment supervision) when tested on SR benchmarks with a maximum of 100 message-passing iterations and early stopping. Given that the SR40-trained model achieved only 3\% SAT accuracy on SR400 (see Table~\ref{tab:mod40onbench}), we focused on evaluating how training on larger instances improves scaling. The results show dramatic improvement on larger benchmarks (36.5\% vs 3\% on SR400), demonstrating that training on larger problems significantly enhances generalization capacity. The "Steps" metrics confirm the SR100-trained model requires fewer iterations on larger problems (e.g., 30.68 vs 41.33 average iterations for SAT instances on SR400).}
\label{tab:mod100onbench}
\begin{tabular}{lccccc}
\hline
Dataset & Decision & SAT Instances & UNSAT Instances & SAT Steps & UNSAT Steps \\
 & Accuracy & Solved & (gap $==$ 1) & (Avg/Med) & (Avg/Med) \\
\hline
SR40 & 90.6\% & 81.2\% & 90.0\% & 14.57/12.0 & 16.21/12.0 \\
SR100 & 79.3\% & 58.7\% & 85.7\% & 22.16/18.0 & 22.64/19.0 \\
SR200 & 83.0\% & 68.2\% & 60.2\% & 22.29/20.0 & 27.12/22.0 \\
SR400 & 68.2\% & 36.5\% & 78.0\% & 30.68/26.0 & 33.13/28.0 \\
\hline
\end{tabular}
\end{table}

A key property of our recurrent GNN architecture for SAT solving is the ability to adjust computational effort at inference time. Unlike standard GNNs with fixed number of layers, the weight-shared recurrent design enables flexible scaling through additional iterations and resampling.

\subsubsection{Iteration and Resampling Effects}

Figure \ref{fig:sat_evolution} demonstrates how increasing message-passing iterations improves the percentage of solved SAT instances. Similarly, Figure \ref{fig:gap_evolution} shows how the average gap decreases across iterations for various benchmarks. The heat maps in Figure \ref{fig:metrics_heatmaps} provide a comprehensive view of how performance metrics improve with both increased iterations and resampling attempts.

For the model trained on SR40, several observations are notable:

\begin{itemize}
    \item \textbf{Iteration benefits}: Increasing iterations from 25 to 125 consistently improves all metrics across benchmarks.
    \item \textbf{Resampling effects}: Multiple inference attempts with different random initializations of node feature vectors further enhance performance. For SR40, decision accuracy improves from 84\% with one sample to 93\% with five samples at 125 iterations.
    \item \textbf{Cross-distribution applicability}: The model trained on SR40 maintains reasonable effectiveness on SR100 and 3SAT100, though with expected performance decrease. This aligns with findings from Li et al. \cite{li2023g4satbench}, who demonstrated that models trained on SR distributions generally transfer well to other SAT problem structures.
\end{itemize}

\subsubsection{Train-time vs Test-time Scaling}

Tables \ref{tab:mod40onbench} and \ref{tab:mod100onbench} present the performance of models trained on SR40 and SR100 distributions when evaluated across benchmarks of varying sizes. The SR40-trained model achieves reasonable generalization to larger instances, though with decreasing effectiveness as problem size increases. For SR100, the model achieves 74.2\% decision accuracy despite being trained on smaller instances, showing good generalization capabilities.

The SR100-trained model demonstrates better performance on larger instances compared to the SR40-trained model, as expected. On SR200, it achieves 83.0\% decision accuracy compared to 58.5\% for the SR40 model. This suggests that while test-time scaling can improve performance on larger problems, there are limits to this approach, and training models on larger instances might be necessary for optimal performance on very large problems.

These results highlight that recurrent GNN architectures allow for a flexible computation-performance tradeoff that can be adjusted at inference time based on available computational resources and desired solution quality.

\subsection{Diffusion Model Extension}\label{sec:diffext}
As we mentioned in Section \ref{sec:diff_schdl_bcgr}, one can use the GNN as a function approximator $f_\theta(\cdot)$ inside a diffusion model. This enables another way of scaling the test-time compute. We adapt the diffusion model used by Sun et al. \cite{sun2023difusco} where the function approximator is trained to predict the ground truth solution $\mathbf{x_0} = f_\theta(\mathbf{x_t},t)$ conditioned on a sample $\mathbf{x_t}$ at time $t$. The predicted assignment is then used to obtain a sample at time $t-1$ and this process is repeated again $\mathbf{x_0} = f_\theta(\mathbf{x_{t-1}},t-1)$ until we reach $t=0$. One application of the function approximator together with the sampling is called a diffusion step. The number of diffusion steps $T$ used for inference is a parameter which can be adapted after the model was already trained and therefore, in this setting we have two types of iterations. One is the number of message-passing iterations and the second is the number of diffusion steps. In Table \ref{tab:gnn_diffusion_perf} we report the tradeoff between the number of message-passing steps (referred to as \texttt{GNN\_Steps}) and the number of diffusion steps. The reported numbers correspond to the dataset \texttt{SR100} with $100$ variables in each problem. The model was trained on the \texttt{SR40} distribution and the tested combinations use around $300$ iterations in total distributed between the two types of steps.

The experiments revealed a consistent trend: increasing the number of message-passing steps is generally more important for improving metrics such as Accuracy and Avg. Gap. 

\subsubsection{Connection to Assignment Prediction Training}
\label{sec:diff_assignment_connection}
We also report an interesting finding which allows to simplify the function approximator used in the diffusion model. Notice that in the expression $\mathbf{x_0} = f_\theta(\mathbf{x_t},t)$ it is also conditioned on the timestep $t$. This conditioning is dictated by the theory of diffusion models \cite{nakkiran2024step} and most of the models, including the one by Sun et al. \cite{sun2023difusco} blindly follow this design choice. In our experiments, we found out that this conditioning is not needed and that the model sometimes works even better without it. Therefore, in all reported results, the model is trained to predict the solution $\mathbf{x_0}$ only from the sample at timestep $t$ ($\mathbf{x_0} = f_\theta(\mathbf{x_t})$). 
Concurrently to us, this fact was also discovered by Sun et al. \cite{sun2025noise} (the same surname is a coincidence) and it's possible that many of the reported experiments which blindly use this conditioning would result in better values without it. 

In this simplified setup, the training examples $(\mathbf{x_0},\mathbf{x_t})$ are sampled by taking a solution of a formula ($\mathbf{x_0}$), sampling a random $t$ from the diffusion schedule and obtaining a corrupted version of the solution at time $t$ ($\mathbf{x_t}$). The model is trained to predict $\mathbf{x_0}$ from $\mathbf{x_t}$. The GNN is the same as in the case of \emph{assignment prediction} except that it also contains a learnable embedding layer which embeds the Boolean values in the assignment $\mathbf{x_t}$ into a vector space to obtain the initial embeddings of variables (or literals) for the first pass of message-passing. 

The only difference from the model trained for \emph{assignment prediction} is therefore that the initial embeddings are not sampled randomly but obtained by embedding the perturbed assignment $\mathbf{x_t}$. This also means that during test time, these two approaches differ only by rounding, i.e. running the model trained for \emph{assignment prediction} for $100$ steps and after every $20$ steps rounding the variable embeddings to vectors representing \emph{True} and \emph{False} is same as running the diffusion model for 5 diffusion steps where each step has 20 message-passing iterations.  

\begin{table}
  \centering
  \caption{Performance Metrics for Different GNN and Diffusion Step Configurations. Variable names shown in parentheses in the original data source are omitted here for brevity.}
  \label{tab:gnn_diffusion_perf} % Use a label for cross-referencing
  % Define number formats for each column using siunitx S columns
  \begin{tabular}{@{} S[table-format=2.0] S[table-format=2.0] S[table-format=1.2] S[table-format=2.1] S[table-format=1.2] S[table-format=2.1] @{}} % @{} removes padding at table edges
    \toprule
    % Multi-row header for clarity
    {\textbf{GNN}} & {\textbf{Diffusion}} & {\textbf{Avg.}} & {\textbf{Dec. Acc.}} & {\textbf{Single-Step}} & {\textbf{Single-Step}} \\
    {\textbf{Steps}} & {\textbf{Steps}} & {\textbf{Gap}} & {\textbf{(\%)}} & {\textbf{Avg. Gap}} & {\textbf{Dec. Acc (\%)}} \\
    \midrule
    20 & 15 & 1.09 & 68.7 & 3.26 & 60.2 \\
    23 & 13 & 1.05 & 70.4 & 2.80 & 63.1 \\ % Added trailing 0 for consistency in S column
    27 & 11 & 0.96 & 73.6 & 2.44 & 64.7 \\
    31 & 9  & 0.98 & 73.8 & 2.11 & 68.2 \\
    35 & 8  & 0.93 & 75.4 & 1.96 & 69.5 \\
    38 & 7  & 0.92 & 76.3 & 1.83 & 70.2 \\
    42 & 7  & 0.90 & 76.7 & 1.70 & 71.6 \\ % Added trailing 0 for consistency in S column
    46 & 6  & 0.95 & 76.8 & 1.64 & 72.1 \\
    50 & 6  & 0.94 & 76.2 & 1.52 & 73.0 \\ % Added trailing 0 for consistency in S column
    \bottomrule
  \end{tabular}
\end{table}

\subsubsection{Interleaving Diffusion Steps with Unit Propagation}
The fact that for each diffusion step, the model outputs probabilities for two possible values, allows us to obtain a partial solution and then run a unit propagation to deduce assignment to other variables. The partial assignment can be obtained by fixing a threshold and then assigning only variables for which one of the values has a predicted probability higher than this threshold. The lower the threshold, the more variables will be fixed and the higher the probability that it will not be possible to complete the partial assignment to a satisfiable assignment. 

We therefore design a tree-search-like algorithm which first tries a low threshold in each diffusion step and if it does not find a satisfiable assignment it backtracks and increases the threshold to obtain a new partial assignment. The details of this algorithm are described in \ref{ap:up} and the experimental results are reported in Table \ref{tab:upp_perf}. As can be seen, interleaving the diffusion steps with unit propagation results in additional improvements over the base diffusion model (approximately $10\%$). We explicitly mention that this experiment is provided only to show a possible avenue for further improvements and the algorithm in its current form is not optimized for speed.

\begin{table}[ht]
  \centering
  \caption{Performance with Unit Propagation. Here we compare the performance with (U.P. Acc.) and without (Dec. Acc.) Unit Propagation, and report the computational cost of Unit Propagation, listing the average number of total recursive function calls, the average number of recursive calls in solved problems, and the average number of recursive calls in unsolved problems.}
  \label{tab:upp_perf}
  \resizebox{\textwidth}{!}{%
  \begin{tabular}{@{}llcccc@{}}
    \toprule
    \textbf{Problems} & \textbf{Dec. Acc. (\%)} & \textbf{U.P. Acc. (\%)} & \textbf{Total Rec. Calls} & 
    \textbf{Solved Rec. Calls} &
    \textbf{Unsolved Rec. Calls}\\ 
    \midrule
    \emph{SR40}  & 88.4 & 94.2 & 32.864 & 6.701 & 53.546\\
    \emph{SR50}  & 86.6 & 93.7 & 29.539 & 6.995 & 47.038\\ 
    \emph{SR60}  & 83.3 & 92.2 & 26.414 & 7.526 & 40.204\\
    \emph{SR70}  & 79.5 & 89.5 & 24.162 & 6.752 & 35.505\\
    \emph{SR80}  & 77.6 & 88.0 & 22.604 & 6.917 & 32.219\\
    \emph{SR90}  & 74.0 & 85.1 & 22.140 & 7.274 & 30.151\\
    \emph{SR100} & 73.4 & 83.7 & 20.363 & 7.129 & 27.074\\
    \emph{SR150} & 63.2 & 75.1 & 17.388 & 7.828 & 20.592\\
    \emph{SR200} & 58.0 & 67.5 & 16.270 & 8.710 & 17.868\\
    \bottomrule
  \end{tabular}%
  }
\end{table}

\section{Interpreting the Trained Model}
\label{sec:interp}
\subsection{Embedding Space Analysis}
Our analysis of variable embeddings reveals patterns that explain how GNNs learn to solve SAT problems. When visualizing these embeddings using dimensionality reduction \cite{mcinnes2018umap}, we observe that they form distinct clusters corresponding to optimal variable assignments.

\begin{figure*}
    \centering
    
    \includegraphics[width=0.9\textwidth]{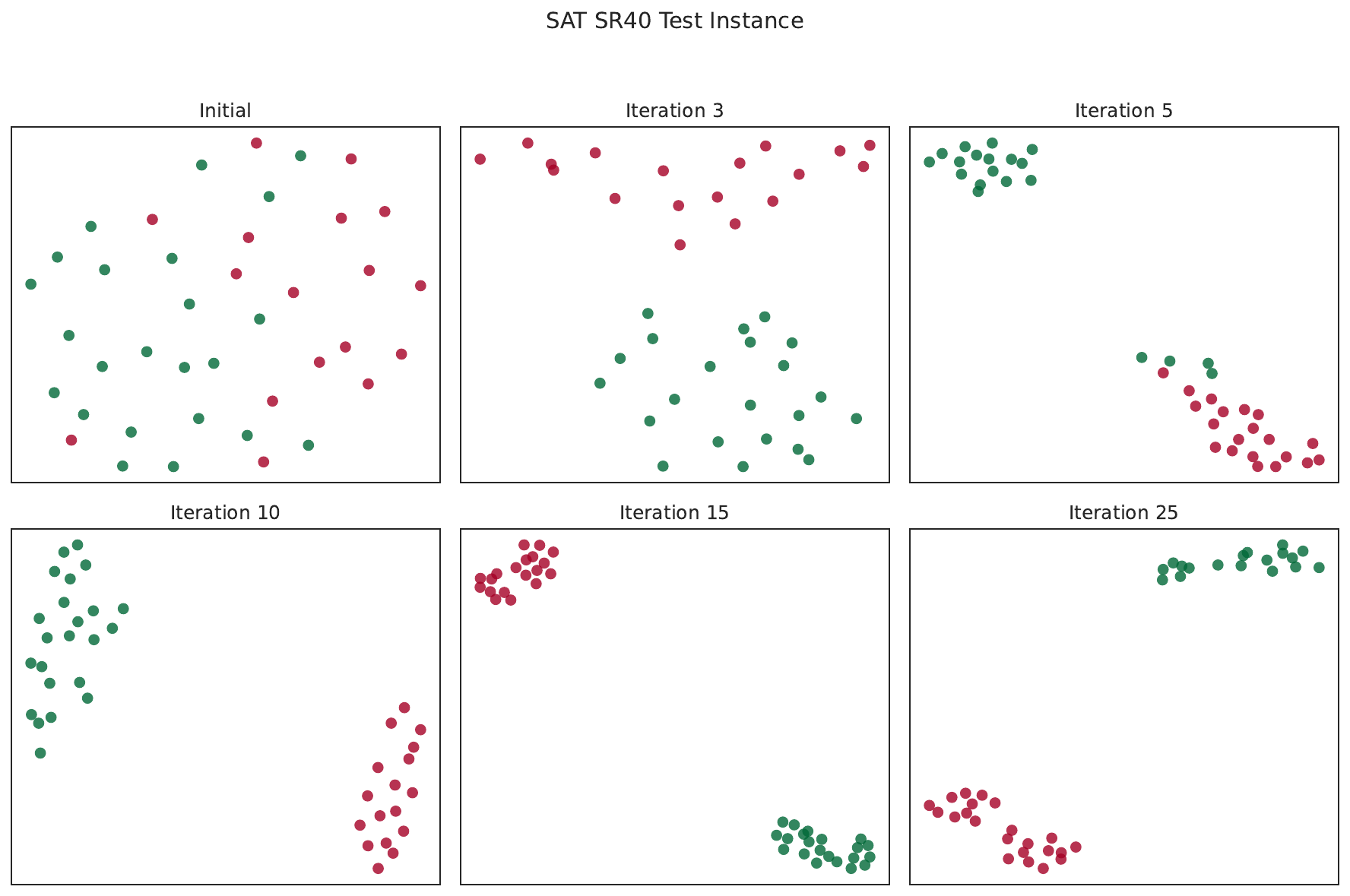}
    \caption{Evolution of variable embeddings during message passing iterations for a satisfiable SR40 instance. The visualization shows 2D projections at different stages (Initial through Iteration 25), colored k-means algorithm in each iteration (green/red). Initially random, embeddings gradually organize into two distinct clusters often corresponding to optimal variable assignments. This clustering behavior was observed across different model architectures and training objectives—notably, even models trained solely for SAT/UNSAT classification (without explicit assignment supervision) develop this embedding separation. This phenomenon supports our interpretation that GNNs implicitly perform continuous optimization similar to SDP relaxation for SAT problems.}
    \label{fig:clustering}
\end{figure*}

As shown in Figure~\ref{fig:clustering}, variable embeddings start randomly distributed but gradually organize into two clusters through message passing iterations. By applying k-means clustering ($k=2$) to these embeddings, we can recover variable assignments that approximate optimal solutions, even from networks trained only to predict satisfiability status.

\subsection{Iterative Optimization Behavior}
By tracking clause satisfaction across iterations, we observe that GNNs solve SAT problems through progressive local refinement. The gap (number of unsatisfied clauses) decreases following a trajectory typical of iterative optimization methods: rapid initial improvement followed by gradual refinement.

This behavior supports the interpretation that GNNs implicitly learn to perform continuous optimization in a high-dimensional space similar to SDP relaxations for SAT. The effectiveness of additional message passing iterations during inference further strengthens this connection. A difference from the SDP relaxation is that the objective function which the GNN implicitely optimizes is non-convex because we observed that it can get stuck in local optima or converge to different solutions when initialized multiple times by different random embeddings.  

Figure \ref{fig:gap_evolution} illustrates how the average gap decreases with increasing iterations. The trajectory suggests a rapid improvement phase followed by more gradual refinement. Individual instance trajectories reveal that while most instances show steady improvement toward optimal solutions, some exhibit fluctuations, particularly unsatisfiable instances. This observation supports the potential value of early stopping techniques, as in rare cases, the gap at later iterations might be higher than a previously achieved minimum gap.

The bi-level optimization perspective—where message passing performs an inner optimization loop (finding variable assignments) guided by network parameters optimized at the outer level (during training)—helps explain the network's ability to generalize to novel problem instances and larger problems than those seen during training. In Section \ref{sec:discuss}, we discuss more details about a possibility of manual derivation of the GNN equations from and explicit objective function. 

\section{Discussion}
\label{sec:discuss}
In this section, we discuss the limitations of our work along with an outlook for future research. The primary limitation of the methods presented here is that they are not competitive with state-of-the-art SAT solvers on benchmarks derived from real-world problems. Current SAT solvers can handle formulas with millions of variables, which is not feasible for the GNN in its current form. However, as mentioned in the introduction, our motivation for studying these models is to better understand the reasoning capabilities of neural networks in a simplified context.

The test-time scaling experiments clearly demonstrate that the GNNs can successfully generalize beyond their training distribution and do not merely learn superficial statistical patterns. The qualitative results presented in Section \ref{sec:interp} further suggest that it is possible to fully understand the mechanisms by which the GNN solves a given formula. Figure \ref{fig:gap_evolution} illustrates that the trained GNN functions as an implicit MaxSAT solver, incrementally maximizing the number of satisfied clauses at each step. These local updates occur in continuous space and can therefore be viewed as gradient updates with respect to an implicit objective function measuring clause satisfaction. Variables are also represented in a high-dimensional vector space, similar to semi-definite programming as explained in \ref{ap:sdp}.

From this perspective, Equations \ref{eq:clause_up}, \ref{eq:var_up}, and \ref{eq:norm} can be interpreted as a gradient descent algorithm searching for an optimal assignment over a high-dimensional unit sphere (due to unit normalization), while the final classification layer corresponds to a rounding step to Boolean values. In future work, we aim to manually derive these equations from a trained GNN using a primal-dual approach, interleaving gradient updates of primal and dual variables associated with constraints. We believe that by utilizing suitable proximal operators and an appropriate metric in the relaxed solution space, the GNN can be effectively interpreted as a primal-dual algorithm optimizing a continuous relaxation of the MaxSAT objective in a high-dimensional space. This points out to another major advantage of using the RNN update function because its simple form is suitable for such derivation.

Deriving equations for such algorithms applicable to arbitrary combinatorial optimization problems would be highly beneficial in practice, allowing these equations to be parameterized by learnable matrices and fine-tuned for specific problem distributions. Such data-driven solvers would be analogous to physics-informed neural networks \cite{cai2021physics}, where substantial domain knowledge is embedded within the model, followed by fine-tuning to approximate the dynamics of a particular physical system. This approach results in fast numerical solvers tailored to specific domains. We believe that the development of data-drive numerical solvers represents an exciting future direction for combinatorial optimization research. To make these numerical solvers practical, it will still be necessary to integrate them into more complex systems, where they would function as guessing or bounding heuristic.

Another limitation of our work is that the model was tested exclusively on random problems. This decision is justified by the findings of Li et al. \cite{li2023g4satbench}, who demonstrated that models trained on random problem instances exhibit superior generalization to other distributions. Since Li et al. already provided experimental results demonstrating the transferability of models across different problem distributions, we chose not to repeat those experiments here.

\section{Conclusion}
\label{sec:concl}
This work provides a comprehensive analysis of graph neural networks for Boolean satisfiability problems. Our evaluation identified key design choices that enhance performance: variable-clause graph representation with RNN updates offers an effective balance of accuracy and efficiency, while our novel closest assignment supervision method significantly improves performance on problems with large solution spaces. The recurrent architecture enables flexible scaling during inference through additional message-passing iterations and resampling. Our diffusion model extension demonstrates another approach to inference-time adaptation, with further improvements possible by integrating classical techniques like unit propagation.

Our analysis of embedding space patterns and optimization trajectories supports the interpretation that these models implicitly implement continuous relaxation algorithms for MaxSAT, explaining their ability to generalize to novel problem instances. This connection provides a theoretical framework for understanding neural reasoning capabilities in structured domains, with implications for the design of hybrid solving approaches.

% References
\bibliographystyle{plainnat}
\bibliography{references}
\newpage
\appendix
\section{Appendix}
\label{sec:appendix}
% Appendix content here
\subsection{Training Tricks and Information}

\subsubsection{Curriculum Learning}\label{curric}
We implement a curriculum learning strategy to improve training efficiency and generalization. The key insight is that starting with simpler (smaller) formulas and gradually increasing complexity allows the model to learn basic logical reasoning patterns before tackling more complex instances.

Our curriculum proceeds as follows:
\begin{enumerate}
    \item Start training with small formulas (5 variables)
    \item Set a validation accuracy threshold for each formula size (starting at 65\% for the smallest size and increasing to 85\% for the largest)
    \item Once the model reaches the threshold accuracy on the current size or reaches a maximum number of epochs (100), increase the formula size by 2 variables
    \item When introducing a new size, include formulas from the four previous sizes to prevent catastrophic forgetting
    \item Continue until reaching the maximum formula size (40 variables)
\end{enumerate}

This curriculum approach significantly accelerates training compared to starting with the full distribution of formula sizes. Our previous experiments showed that the curriculum-trained model reaches 85\% validation accuracy in approximately 30 minutes, compared to over 5 hours for the non-curriculum approach.

\begin{figure*}
    \centering
    \includegraphics[width=0.9\textwidth]{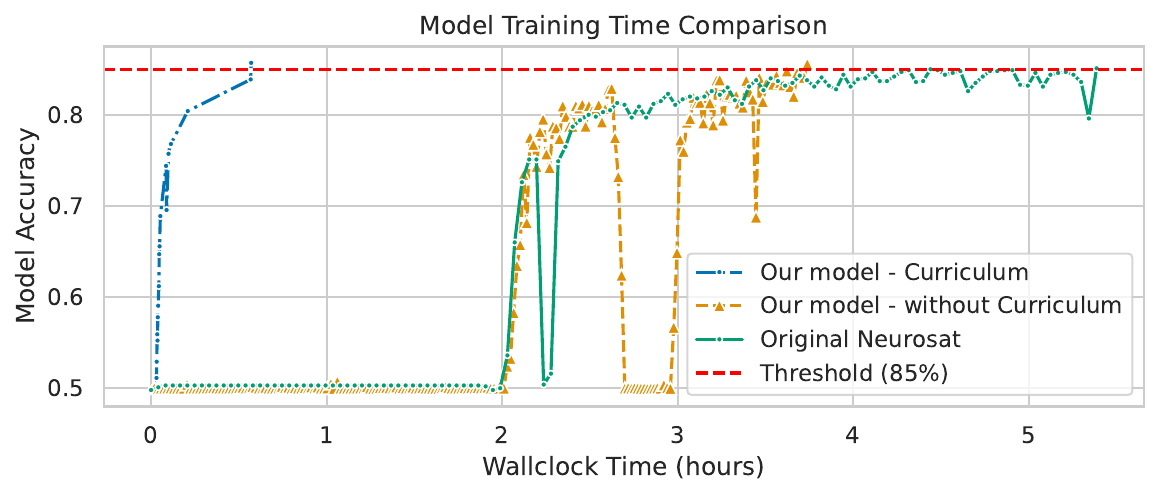}
    \caption{Validation accuracy during training. Our model with a curriculum achieves reaches 85\% in approximately 30 minutes, whereas the original NeuroSAT implementation needs over 5 hours. For comparison, we also add our implementation trained on the same data, but without a curriculum. The training of each model stops once it achieves an accuracy of 85\% on a validation set.}
\vspace{10pt}
    \label{fig:model_performance}
\end{figure*} 

\subsubsection{Exponential Moving Average (EMA)}\label{aema}
We employ Exponential Moving Average (EMA) for model parameter updates during training. EMA maintains a shadow copy of the model parameters that is updated after each training batch:
\begin{align}
\theta_{\text{EMA}} \leftarrow \beta \theta_{\text{EMA}} + (1-\beta) \theta_{\text{current}}
\end{align}
where $\beta$ is the decay rate (we use $\beta = 0.999$).

During validation and testing, we use the EMA parameters instead of the current parameters. This technique significantly stabilizes training and improves generalization, especially in the early stages of training. Our experiments show that EMA provides a smooth validation accuracy curve, while the validation accuracy of the non-EMA model exhibits high variance and jumps of up to 10\%.

\subsubsection{Learning Rate Schedule}
We implement a custom learning rate schedule that combines cosine annealing for the first half of training and a constant minimum learning rate for the second half:
\begin{align}
\eta(t) = 
\begin{cases}
\eta_{\min} + (\eta_0 - \eta_{\min})\frac{1 + \cos(\pi t / t_{\text{half}})}{2} & \text{if } t < t_{\text{half}} \\
\eta_{\min} & \text{otherwise}
\end{cases}
\end{align}
where $\eta_0$ is the initial learning rate, $\eta_{\min}$ is the minimum learning rate (set to $10^{-5}$), $t$ is the current epoch, and $t_{\text{half}}$ is half of the maximum number of epochs.

This schedule helps the model converge to a good solution in the first half of training and then fine-tune in the second half without disrupting the learned representations.

\subsubsection{Impact of hidden dimension on GNN Performance}\label{appdim}

The dimensionality of the hidden representations, here denoted as \texttt{d\_model}, specifies the size of the embedding vectors of variables and the hidden state dimension used during the message passing and update phases within the GNN architecture.

The choice of \texttt{d\_model} directly influences the model's capacity to learn complex patterns and relationships within the graph structure and node features. It also impacts computational resource requirements, such as memory usage and training time. Understanding how performance metrics vary with different \texttt{d\_model} values is therefore crucial for effective model design and hyperparameter tuning. 

Our evaluation in table \ref{tab:gnn_emb_size_impact} generally shows that increasing the \texttt{d\_model} leads to improved model performance, likely due to the enhanced representational capacity allowing the model to capture more intricate features. However, we observed that this trend exhibits diminishing returns; while significant performance gains are noticeable as the dimension increases up to 64, further increases yield smaller improvements in accuracy relative to the growing computational cost (e.g., peak accuracy at \texttt{d\_model}=256 came with significantly longer training time). This suggests that, considering the marginal benefits, a dimension around 64 still presents a practical optimum, offering a good balance between performance and model complexity/efficiency for this specific setup.

\begin{table} [htbp]
\centering  
\caption{Experimental results demonstrating the impact of hidden dimension size (\texttt{d\_model}) on model performance and training duration. 
`Embedding Size (\texttt{d\_model})` refers to the dimensionality of the hidden representations within the GNN. 
`Accuracy` indicates the performance metric achieved by the model. 
`Time (hours)` specifies the total time required to train the model for each corresponding dimension size.}
\label{tab:gnn_emb_size_impact}  
\begin{tabular}{ccc}
\hline
Embedding Size (\texttt{d\_model}) & Accuracy & Time (hours) \\
\hline
16  & 0.782 & 1.62 \\
32  & 0.852 & 1.66 \\
48  & 0.860 & 1.70 \\
64  & 0.869 & 1.87 \\
96  & 0.861 & 2.32 \\
128 & 0.864 & 2.99 \\
256 & 0.877 & 7.55 \\
\hline
\end{tabular}

\end{table}

\subsection{Diffusion Model Extensions}

\subsection{Using Unit Propagation for Problem Simplification in the Diffusion Process}
\label{ap:up}
In the diffusion process, each iteration provides a belief for every variable, which can be leveraged to continuously simplify the problem via a unit propagation algorithm until it converges to an empty problem, thereby obtaining a solution. The overall solving process is recursive, and its main steps are described as follows:

\begin{enumerate}
    \item \textbf{Partial Assignment Extraction and Local Unit Propagation} \\
    In each diffusion step, a belief value between 0 and 1 is calculated for every variable. A value closer to 1 indicates a stronger inclination toward being \textit{true}, and vice versa. We set a threshold to select variables with high belief and assign them accordingly to obtain a partial assignment. This partial assignment is then used to perform unit propagation for clause simplification. The unit propagation algorithm works as follows: \begin{itemize}
        \item If a clause contains a literal that is already satisfied by the current assignment, the clause is marked as satisfied.
        \item If all literals in a clause have been assigned but none satisfy it, a conflict signal is returned.
        \item For clauses that are not fully assigned, the unassigned literals are retained to form a simplified clause set.
        \item For unit clauses obtained during the simplification process (i.e., clauses containing only a single literal), the corresponding unassigned variable is directly assigned the appropriate value, further advancing the local solving process.
    \end{itemize}

    \item \textbf{Multi-Threshold Strategy and Recursive Solving} \\
    In the partial assignment extraction step, setting a lower threshold allows for the selection of as many assignments as possible at each step, thereby greatly simplifying the problem; however, it is more likely to select unreliable assignments that may lead to contradictions. To balance this, we adopt a multi-threshold list, starting from the lowest threshold. For each given threshold, if a new partial assignment is obtained, unit propagation is used to update the clauses and evaluate:
    \begin{itemize}
        \item If the simplified clause set becomes empty, all clauses are satisfied and the final solution is directly returned.
        \item If a conflict occurs or the recursive call at the next level fails, the threshold is raised; if the highest threshold is reached, the process moves to the next recursive level and performs another diffusion step.
        \item If unit propagation succeeds but the problem is not yet completely solved, the process recurses to the next level, performing a diffusion step on the updated clauses.
        \item If the recursion reaches a preset maximum depth and the clauses still cannot be satisfied, the recursion at that level fails.
    \end{itemize}
\end{enumerate}

Table~\ref{tab:upp_perf} shows the performance and computational cost after applying unit propagation. We tested a fixed model under the settings: GNN steps = 25, diffusion steps = 10, and multi-threshold list = [0.6, 0.75, 0.9]. As shown, incorporating unit propagation improves accuracy by approximately 10\% across various problem settings. The last three columns of the table list the number of recursive function calls during the recursion process. Since each call involves one diffusion step, the computational cost incurred by the multi-threshold strategy is directly reflected. We observed that for harder problems, the computational cost of the multi-threshold strategy is actually lower, as unit propagation on partial assignments is more likely to encounter conflicts, thereby reducing the number of recursive branches.

\subsection{Influence of Number of Message-passing and Diffusion Steps}
For completeness, we also report evaluations in which the number of diffusion steps is fixed and the number of message-passing steps is changing (Table \ref{tab:gnn_increasing_perf}) and vice versa (Table \ref{tab:diffusion_increasing_perf}). We observe that the expansion of the number of iterative steps does not always bring benefits: when the number of one kind of step is fixed, further increasing the number of another kind of step beyond a certain threshold will not lead to performance improvement.

\begin{table}
  \centering
  \caption{Performance on SR100 for Different Diffusion Step and Fixed GNN Step. Note that the Performance is no longer Significantly Improved when Diffusion Steps Larger than 8.}
  \label{tab:diffusion_increasing_perf}
  \begin{tabular}{@{} S[table-format=2.0] S[table-format=2.0] S[table-format=1.2] S[table-format=2.1] @{}} % @{} removes padding at table edges
    \toprule
    % Multi-row header for clarity
    {\textbf{GNN Steps}} & {\textbf{Diffusion Steps}} & {\textbf{Avg. Gap}} & {\textbf{Accuracy (\%)}} \\ 
    \midrule
    25 & 4  & 0.991 & 69.9 \\
    25 & 5  & 0.901 & 71.2 \\
    25 & 6  & 0.798 & 72.7 \\ 
    25 & 8  & 0.705 & 73.0 \\
    25 & 10 & 0.728 & 72.1 \\
    25 & 20 & 0.662 & 73.3 \\ 
    25 & 30 & 0.676 & 72.3 \\
    25 & 40 & 0.655 & 73.3 \\
    25 & 50 & 0.663 & 73.0 \\
    \bottomrule
  \end{tabular}
\end{table}

\begin{table}
  \centering
  \caption{Performance on SR100 for Different GNN Step and Fixed Diffusion Step. Note that the Performance is no longer Significantly Improved when GNN Steps Larger than 50.}
  \label{tab:gnn_increasing_perf}
  \begin{tabular}{@{} S[table-format=2.0] S[table-format=2.0] S[table-format=1.2] S[table-format=2.1] @{}} % @{} removes padding at table edges
    \toprule
    % Multi-row header for clarity
    {\textbf{GNN Steps}} & {\textbf{Diffusion Steps}} & {\textbf{Avg. Gap}} & {\textbf{Accuracy (\%)}} \\ 
    \midrule
    10 & 10 & 2.028 & 55.0 \\
    20 & 10 & 0.846 & 68.6 \\ 
    30 & 10 & 0.622 & 74.4 \\
    40 & 10 & 0.578 & 75.5 \\
    50 & 10 & 0.533 & 77.6 \\
    60 & 10 & 0.518 & 77.2 \\
    70 & 10 & 0.500 & 78.6 \\
    80 & 10 & 0.521 & 77.9 \\
    90 & 10 & 0.512 & 77.6 \\
    100& 10 & 0.522 & 77.4 \\
    \bottomrule
  \end{tabular}
\end{table}

\section{SDP for MAX-2-SAT} \label{ap:sdp}

Semidefinite programming (SDP) is a mathematical optimization technique primarily used for problems involving positive semidefinite matrices. In SDP, a linear objective function is optimized over a feasible region given by a \emph{spectrahedron} (an intersection of a convex cone formed by positive semidefinite matrices and an affine subspace) \cite{ramana1995some}. Along with the broad scope of applications, SDP has been used to design approximation algorithms for discrete NP-hard problems \cite{gartner2012approximation}. This is achieved by lifting variables of a problem to a vector space and optimizing a loss function expressed in terms of these vectors.

In this section, we provide a detailed derivation of the SDP relaxation for MAX-2-SAT. The goal is to write an objective function for 2-CNF formulae, which consist of clauses $c_1, \dots, c_k$ over variables $x_1, \dots, x_n$ with at most two literals per clause.

\subsection{Derivation of the SDP Relaxation}

For each Boolean variable $x_i$ (where $i \in \{1, 2, \ldots, n\}$), a new variable $y_i \in \{-1,1\}$ is associated, and an additional variable $y_0 \in \{-1,1\}$ is introduced. This additional variable is introduced to unambiguously assign the truth value in the original problem from values of the relaxed problem. It is not possible to just assign \emph{True} (\emph{False}) to $x_i$ if $y_i = 1 (-1)$ because quadratic terms cannot distinguish between $y_i \cdot y_j$ and $(-y_i) \cdot (-y_j)$. Instead, the truth value of $x_i$ is assigned by comparing $y_i$ with $y_0$: $x_i$ is \emph{True} if and only if $y_i = y_0$, otherwise it is \emph{False}. The assignment is therefore invariant to negating all variables.

To determine the value of a formula, we sum the value of its clauses $c$ which are given by the value function $v(c)$. Here are examples of the value function for different clauses:

\begin{align}
v(x_i) &= \frac{1+y_0 \cdot y_i}{2} \\
v(\neg x_i) &= 1 - v(x_i) = \frac{1-y_0 \cdot y_i}{2} \\
v(x_i \vee \neg x_j) &= 1 - v(\neg x_i \wedge x_j) \\
&= 1 - \frac{1-y_0 \cdot y_i}{2} \cdot \frac{1+y_0 \cdot y_j}{2} \\
&= \frac{1}{4}(1+y_0 \cdot y_i) + \frac{1}{4}(1-y_0 \cdot y_j) + \frac{1}{4}(1 + y_i \cdot y_j)
\end{align}

By summing over all clauses $c$ in the Boolean formula, the following integer quadratic program for MAX-2-SAT is obtained:
\begin{align}
\text{Maximize:} \quad & \sum_{c \in C} v(c) \\
\text{Subject to:} \quad & y_{i} \in \{-1, 1\} \text{ for all } i \in \{0, 1, \ldots, n\}
\end{align}

This can be rewritten by collecting coefficients of $y_i \cdot y_j$ for $i,j \in \{0, 1, \ldots, n\}$ and putting them symmetrically into a $(n+1) \times (n + 1)$ coefficient matrix $W$. The terms $y_i \cdot y_j$ can be collected in a matrix $Y$ with the same dimensions as $W$. The elements $Y_{ij}$ correspond to $y_i \cdot y_j$ for $i,j \in \{0, 1, \ldots, n\}$. Both matrices are symmetric, hence the sum of all elements in their element-wise product (which is the objective function) can be compactly expressed by using the trace operation. This leads to the following version of the same integer program:

\begin{align}
\text{Maximize:} \quad & \text{Tr}(WY) \\
\text{Subject to:} \quad & Y_{ii} = 1 \text{ for all } i \in \{0, 1, \ldots, n\} \\
& Y_{ij} = y_i \cdot y_j \text{ for all } i,j \in \{0, 1, \ldots, n\} \\
& y_i \in \{-1, 1\} \text{ for all } i \in \{0, 1, \ldots, n\}
\end{align}

\subsection{Relaxation to Semidefinite Programming}

\begin{figure}[htbp]
  \centering
  \begin{minipage}[b]{0.3\textwidth}
    \centering
    \includegraphics[width=\linewidth]{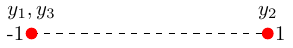}
  \end{minipage}
  \hfill
  \begin{minipage}[b]{0.3\textwidth}
    \centering
    \includegraphics[width=\linewidth]{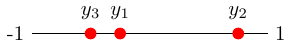}
  \end{minipage}
  \hfill
  \begin{minipage}[b]{0.3\textwidth}
    \centering
    \includegraphics[width=\linewidth]{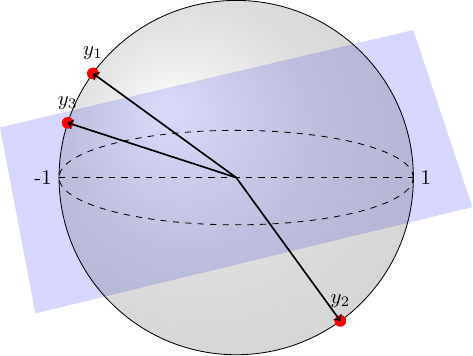}
  \end{minipage}
  \caption{Lifting the variables to a higher dimension, demonstrated on variables $y_1, y_2, y_3$. Initially, only integer values of $-1$ and $1$ could be assigned to them (integer program). Next, constraints are relaxed, allowing variables to take any real value between $-1$ and $1$. Finally, it is permitted for them to be unit vectors in a high-dimensional space (here, 3 dimensions). The hyperplane in the last picture would be used for rounding the variables at the end. This hyperplane can be randomly selected, and truth values for variables $y_1, y_2, y_3$ are determined based on which side of the hyperplane they land after continuous optimization.}
\label{fig:lifting}
\end{figure}

To make the discrete program continuous, we first allow the value of the variables $y_i$ to be any real number between $-1$ and $1$. However, semidefinite programming goes further and allows variables to be $(n+1)$-dimensional unit vectors $({y}_0, \ldots, {y}_n) \longrightarrow (\mathbf{y}_0, \ldots, \mathbf{y}_n)$, as schematically depicted in Figure \ref{fig:lifting}. In this relaxation, the binary products $y_i \cdot y_j$ in the objective function are replaced by inner products $\langle \mathbf{y_i}, \mathbf{y_j} \rangle$.

This can be compactly represented in matrix form by substituting each inner product $\langle \mathbf{y_i}, \mathbf{y_j} \rangle$ with a scalar $Y_{ij}$ of a matrix $Y$. The fact that these scalars correspond to inner products is encoded by the restriction to \emph{positive-semidefinite} matrices $Y \succeq 0$. The SDP relaxation of MAX-2-SAT can thus be formulated as:

\begin{align}
\text{Maximize:} \quad & \text{Tr}(WY) \\
\text{Subject to:} \quad & Y_{ii} = 1 \text{ for all } i \in \{0, 1, \ldots, n\} \\
& Y \succeq 0
\end{align}

Positive semidefiniteness of matrix $Y$ ensures that it can be uniquely factorized as $Y = Y^{\frac{1}{2}} (Y^{\frac{1}{2}})^T$. We can then obtain real unit vectors $\mathbf{y}_i$ for all $i \in \{0, \ldots, n\}$ such that $Y_{ij} = \langle \mathbf{y_i}, \mathbf{y_j} \rangle$ for all $i, j \in \{0, \ldots, n\}$. The constraints $Y_{ii} = 1$ ensure that all vectors $\mathbf{y}_i$ lie on an $(n+1)$-dimensional unit sphere.

\subsection{Interpretation and Rounding}

The SDP solver optimizes the numbers in the matrix $Y$, but using the factorization, we can visualize what happens with the vectors $\mathbf{y_i}$. The process starts with random unit vectors that are continuously updated to maximize the objective function. If we fix the position of the vector $\mathbf{y_0}$ (corresponding to the value \emph{true}), we would see that the vectors of variables that will be set to true in the final assignment get closer to the vector $\mathbf{y_0}$, while the vectors $\mathbf{y_j}$ of variables that will be set to false move away from it so that the inner product $\langle \mathbf{y_0}, \mathbf{y_j} \rangle$ is close to $-1$.

If the formula is satisfiable, the objective function drives the vectors to form two well-separated clusters. However, if only a few clauses can be satisfied simultaneously, the vectors would end up being scattered.

A simple way to round the resulting vectors ($\mathbf{y}_1, \ldots, \mathbf{y}_n$) and get the assignment for the original Boolean variables is to compute the inner product $\langle \mathbf{y_0}, \mathbf{y_i} \rangle$ and assign the value according to its sign. It is also possible to assign the values by picking a random separating hyperplane, and it can be shown that this rounding gives a $0.8785$-approximation of the integer program optimum \cite{goemans1995improved}.

Note that the expressions of the clauses reach their maximum at 1 (when a clause is satisfied by the assignment). This means that the whole formula is satisfiable if the objective function achieves a value equal to the number of clauses in the formula. Another way to check satisfiability is to plug the obtained solution into the formula and verify whether it is satisfied. Therefore, we can obtain an incomplete SAT solver from this SDP formulation.

Similar SDPs can be obtained for different versions of MAX-SAT (with larger clauses). From empirical observation, the convergence threshold of the SDP solver needs to be decreased significantly compared to MAX-2-SAT in order to obtain a good approximation for these more complicated versions.
\end{document}